\documentclass[runningheads]{llncs}
\pdfoutput=1

\usepackage{eccv}

\usepackage{eccvabbrv}

\usepackage{wrapfig}
\usepackage{graphicx}
\usepackage{booktabs}
\usepackage{xcolor}
\usepackage[table]{xcolor}
\usepackage{amsmath}
\usepackage{amssymb}
\usepackage{mathtools}
\usepackage{CJKutf8}
\usepackage{multirow}
\usepackage{enumitem} 
\usepackage[breaklinks,colorlinks,citecolor=eccvblue]{hyperref}
\usepackage{tcolorbox}

\usepackage[accsupp]{axessibility}  

\usepackage{orcidlink}

\newcommand{\equalcontrib}{\ensuremath{^*}}
\newcommand{\internship}{\ensuremath{^\dagger}}
\newcommand{\corresponding}{\ensuremath{^\ddagger}}
\newcommand{\firstauthoremail}{libozhou@pku.edu.cn}
\newcommand{\corrauthoremail}{longo11070001@gmail.com}
\newcommand{\titlefootnotes}{%
  \begingroup
  \renewcommand{\thefootnote}{\fnsymbol{footnote}}%
  \footnotetext[1]{Equal contribution.}%
  \footnotetext[4]{Work done during an internship at Kling Team.}%
  \footnotetext[5]{Corresponding author.}%
  \endgroup
}

\begin{document}

\title{ 
    GRAN-TED: Generating Robust, Aligned, and Nuanced Text Embedding for Diffusion Models} 

\titlerunning{GRAN-TED}

\author{\begin{tabular}{@{}c@{}}
Bozhou Li\inst{1,2}\equalcontrib\internship \quad
Sihan Yang\inst{1,3}\equalcontrib \quad
Yushuo Guan\inst{2}\equalcontrib\\[0.2em]
Ruichuan An\inst{1} \quad
Xinlong Chen\inst{4} \quad
Yang Shi\inst{1}\\[0.2em]
Pengfei Wan\inst{2} \quad
Wentao Zhang\inst{1} \quad
Yuanxing Zhang\inst{2}\corresponding
\end{tabular}}

\authorrunning{B.~Li et al.}

\institute{$^{1}$Peking University \quad
$^{2}$Kling Team, Kuaishou Technology\\
$^{3}$Xi'an Jiaotong University \quad
$^{4}$School of Artificial Intelligence, UCAS\\[-0.1em]
\email{\firstauthoremail} \quad \email{\corrauthoremail}}

\maketitle
\titlefootnotes

\begin{abstract}
The text encoder is a critical component of text-to-image and text-to-video diffusion models, fundamentally determining the semantic fidelity of the generated content. However, its development has been hindered by two major challenges: the lack of an efficient evaluation framework that reliably predicts downstream generation performance, and the difficulty of effectively adapting pretrained language models for visual synthesis. To address these issues, we introduce GRAN-TED, a paradigm to Generate Robust, Aligned, and Nuanced Text Embeddings for Diffusion models. Our contribution is twofold. First, we propose TED-6K, a novel text-only benchmark that enables efficient and robust assessment of an encoder's representational quality without requiring costly end-to-end model training. We demonstrate that performance on TED-6K, standardized via a lightweight, unified adapter, strongly correlates with an encoder's effectiveness in downstream generation tasks. Notably, under our experimental setup, compared with training a diffusion model from scratch, evaluating with TED-6K is about \textbf{750$\times$ faster}. Second, guided by this validated framework, we develop a superior text encoder using a novel two-stage training paradigm. This process involves an initial fine-tuning stage on a Multimodal Large Language Model for better visual representation, followed by a layer-wise weighting method to extract more nuanced and potent text features. Our experiments show that the resulting GRAN-TED encoder not only achieves state-of-the-art performance on TED-6K but also leads to demonstrable performance gains in text-to-image and text-to-video generation. 
\end{abstract}
\section{Introduction}
\label{sec:intro}

Diffusion models have recently emerged as the dominant paradigm for text-to-image (T2I)~\cite{ramesh2021zero,rombach2022high,labs2025flux1kontextflowmatching,gong2025seedream,qin2025lumina,an2025unictokens} and text-to-video (T2V)~\cite{wan2025wan,kong2024hunyuanvideo,wiedemer2025video,ma2025step,gao2025seedance} generation.
Central to these models is the text encoder, which transforms textual prompts into semantic representations that guide the visual synthesis process.
An effective text encoder must accurately represent a wide range of semantic elements, including subject features, static attributes, spatial relationships, and temporal events.
This capability is crucial for enabling the generative model to achieve compositional generalization—synthesizing novel scenes from combinations of concepts not explicitly seen during training. 
Despite its critical role, the importance of the text encoder was largely overlooked in the early stages of diffusion model development, with most research focusing on the diffusion mechanism itself. This oversight is consequential, as the quality of the text encoder fundamentally determines the semantic fidelity between the prompt and the generated visual content.

This trend has begun to reverse, with a clear architectural evolution in text encoders. The community has progressed from using early CLIP-based encoders~\cite{rombach2022high} to more powerful models like T5~\cite{labs2025flux1kontextflowmatching}, and has now converged on integrating Large Language Models (LLMs) as the state-of-the-art backbone~\cite{gong2025seedream,wang2025comprehensive}. The rationale is clear: the advanced semantic understanding of LLMs promises to directly translate into higher-fidelity visual generation, improving the alignment between text and image.

However, despite the adoption of these powerful LLMs, generative models still exhibit significant challenges in prompt fidelity. Common failures observed in practical applications and on benchmark evaluations include incorrect object counts, misinterpretation of relational or referential phrases, attribute binding errors, and a failure to adhere to negative constraints.
As the bridge between textual descriptions and visual synthesis, a more accurate and robust text representation is paramount to addressing these shortcomings.
Developing a superior text encoder tailored for generative models presents two primary challenges:
\begin{itemize}
\item \textbf{Evaluating Representation Quality for Generation}:
A significant hurdle is the lack of a suitable evaluation framework. Standard NLP benchmarks do not correlate directly with visual generation. Furthermore, information retrieval metrics not only miss compositional complexity and nuanced world knowledge, but their single-vector paradigm inherently misaligns with how contemporary diffusion models utilize text embeddings. Conversely, evaluating an encoder via end-to-end training of a full generative model is computationally prohibitive and inefficient.
\item \textbf{Adapting LLM Representations for Visual Synthesis}:
The second challenge lies in effectively adapting the rich features of a pre-trained LLM for the specific demands of visual generation.
It is non-trivial to ensure that the semantic information encoded in the LLM's latent space is represented in a manner that is precise, unambiguous, and fully interpretable by the diffusion model, thereby translating improved textual understanding into tangible gains in generation fidelity.
\end{itemize}

To address these two challenges, we propose GRAN-TED to Generate Robust, Aligned, and Nuanced Text Embedding for Diffusion models. We first introduce a robust and unified evaluation framework centered on a novel text-only benchmark named TED-6K. The TED-6K benchmark, composed of visual captions and verified true/false statements, allows for efficient and robust assessment of an encoder's representational quality. We demonstrate that an encoder's performance on TED-6K, which is measured by representation similarity and standardized via a lightweight adapter, strongly correlates with its effectiveness in downstream text-to-image and text-to-video generation. In our experimental setup , training a T2I diffusion model from scratch for evaluation takes about 50 hours, whereas evaluating an encoder with TED-6K takes only about 4 minutes, enabling \textbf{~750× faster} iteration.

Leveraging this validated framework, we then explore a two-stage paradigm to develop a superior text encoder GRAN-TED. In the first stage, we finetune Qwen3-VL-8B-Instruct~\cite{qwen3vl2025} for better aligned latent visual representation towards generation. In the second stage, we implement a novel layer-wise weighting method, which extracts more nuanced text features from the layers of the trained model in the first stage. Our experiments confirm that this two-stage paradigm yields a new robust text encoder on TED-6K, simultaneously leading to demonstrable performance gains in T2I and T2V models.

\section{Related Works}

\subsection{Text Encoder for Diffusion Models.}

Text encoders are a central conditioning component in text-to-image and
text-to-video diffusion models, yet their design has followed several different
routes. Early systems commonly adopt the CLIP text encoder~\cite{radford2021learning}
or T5-family encoders~\cite{raffel2020exploring}, whose pre-training objectives
provide transferable semantic representations for visual synthesis~\cite{rombach2022high,podell2023sdxl}.
More recent models increasingly replace or augment these encoders with
decoder-only LLMs~\cite{yang2025qwen3,luo2024llm} and
MLLMs~\cite{qwen3vl2025,an2024mc,lin2024draw}, motivated by their stronger
instruction following, compositional understanding, and visually grounded
semantics~\cite{ma2024exploring,li2025ldgen,wan2025wan,lin2025perceiveanythingrecognizeexplain}.

Another line of work improves how language features are injected into diffusion
models. ELLA~\cite{hu2024ellaequipdiffusionmodels}, Deep Fusion~\cite{tang2025exploring},
and Playground v3~\cite{liu2024playground} introduce stronger LLM-based
conditioning paths to enhance prompt alignment, while Seedream~\cite{gong2025seedream}
integrates LLM features with a glyph-aligned model for bilingual image
generation. These methods demonstrate the value of richer language
representations, but they mainly study how to integrate a chosen text encoder
into a generative system. In contrast, our focus is the upstream problem of
evaluating, selecting, and adapting text encoders themselves before costly
end-to-end diffusion training.

Feature extraction from a text encoder is also non-trivial. Although many
diffusion models use hidden states from a single deep layer, recent
research~\cite{dar2022analyzing,skean2024does,wang2025comprehensive} shows
that normalizing and averaging the hidden states from all layers (Norm-Avg)
yields stronger text conditioning than using a single layer or simple averaging:
\begin{equation}
    c_{\text{text}} = \frac{1}{N} \sum_{i=1}^{N} \text{LayerNorm}(h_i).
\end{equation}
Post-training methods further refine the text encoder after a diffusion model
has already been trained~\cite{li2024textcraftor,chen2024enhancing}, and
efficiency-oriented work studies how to reduce the cost of text encoders in
text-to-image systems~\cite{wang2025scaling}. These directions are complementary
to ours. GRAN-TED does not aim to compress the denoising backbone, accelerate
the sampler, or reduce the text encoder to a smaller model. Instead, we ask
which text representations are most predictive of downstream generation quality,
how to evaluate them efficiently in a text-only manner, and how to fuse
layer-wise features into a stable conditioning signal for DiT training.

\subsection{Text-alignment Benchmarks for Text Encoders.}

Prevailing text-alignment benchmarks focus on assessing the correspondence
between the final visual output and the input text prompt. For instance,
GenAI-Bench~\cite{li2024genai,lin2024evaluating} leverages a Vision-Language
Model (VLM) to judge the alignment of generated images in terms of various
attributes. Similarly, VIEScore~\cite{ku2023viescore}, GPT4-Eval~\cite{zhang2023gpt},
and recent video-oriented evaluations~\cite{guo2025video,liu2025improving}
employ powerful multimodal models or reward models to directly score the
generated visual output. These benchmarks are valuable for final model
evaluation, but they assess the diffusion model as a whole rather than isolating
the text encoder's contribution. As a result, using them to compare candidate
encoders requires training or fine-tuning a separate generative model for each
encoder, which is prohibitively expensive.

Some works instead use existing retrieval benchmarks and LLM benchmarks as
proxy evaluations. Retrieval-based benchmarks such as MTEB~\cite{muennighoff2022mteb}
mainly evaluate a single pooled sentence vector, creating a granularity mismatch
with diffusion models that condition on full token sequences~\cite{qin2025lumina,ma2025step,kong2024hunyuanvideo,wan2025wan,gong2025seedream,gao2025seedance}.
General LLM benchmarks~\cite{hendrycks2020measuring,AIME2025} are also
insufficient proxies: their scores reflect answer generation, reasoning, and
decoding ability, whereas a diffusion model consumes static hidden
representations from the text encoder. TED-6K is therefore positioned as a
diagnostic benchmark for text encoders rather than a replacement for
output-level image or video evaluation. It measures whether the encoder
representation preserves fine-grained, visually relevant semantics and whether
this representation-level score predicts downstream generation performance.

\begin{figure*}[t]
    \centering
    \includegraphics[width=1\textwidth]{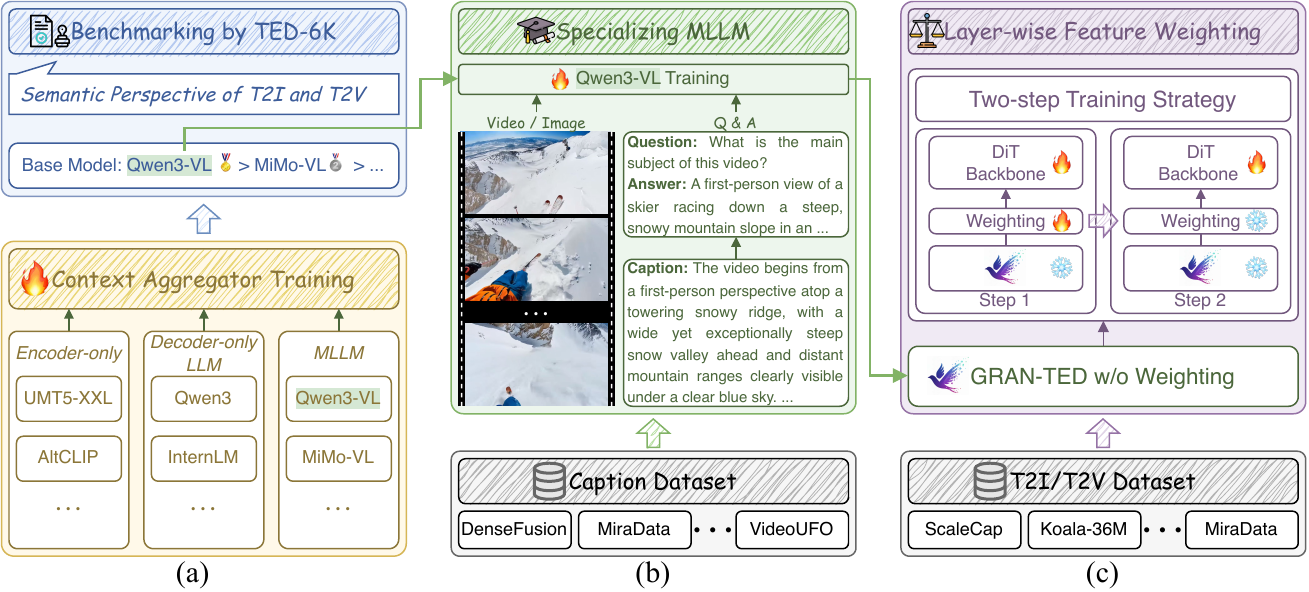}
    \caption{An overview of our complete framework, integrating our evaluation and development pipelines. Figure (a) illustrates the TED Evaluation Framework, consisting of the TED-6K benchmark and a context aggregator to assess the representational capabilities of text encoders. Figure (b) shows the construction of our TED Encoder, where Qwen3-VL-8B-Instruct is fine-tuned on a curated VQA and captioning dataset to specialize the MLLM. Figure (c) depicts our final GRAN-TED solution, which incorperates a learnable layer-wise weighting module to generate GRAN-TED for diffusion models.}
    \label{fig:overview}
\end{figure*}

\section{Method}

\subsection{Overview}

As illustrated in ~\cref{fig:overview}, we have developed a robust pipeline designed to identify the optimal text encoder for diffusion models from a vast pool of candidates.
First, we introduce TED-6K, a benchmark specifically tailored to the domain of visual generation. Unlike traditional methods~\cite{lin2024evaluating,ku2023viescore,guo2025video}, TED-6K serves as a text-only evaluation benchmark, enabling a multidimensional assessment of an encoder's representational capabilities without the prohibitive cost of training a full diffusion model end-to-end.
Next, we develop a unified context aggregator to ensure a fair comparison across diverse architectures, including CLIP, T5, LLMs and MLLMs, which allows us to equitably evaluate the representational quality of various models on the TED-6K.
Guided by this rigorous screening process, we identify Qwen3-VL-8B-Instruct~\cite{qwen3vl2025} as the superior backbone. Building upon this backbone, we fine-tune it on the VQA dataset to develop an enhanced text encoder, termed GRAN-TED.
Furthermore, we investigate strategies to efficiently fuse distinct feature layers within GRAN-TED, thereby maximizing the quality of the resulting text embeddings. The specifics of the benchmark formulation, the context aggregator architecture, our developed encoder, and the feature fusion strategy are detailed in  \cref{sec:text-only benchmark} - \cref{sec:feature combi}, respectively.

\begin{figure*}[t]
    \centering
    \includegraphics[width=\textwidth]{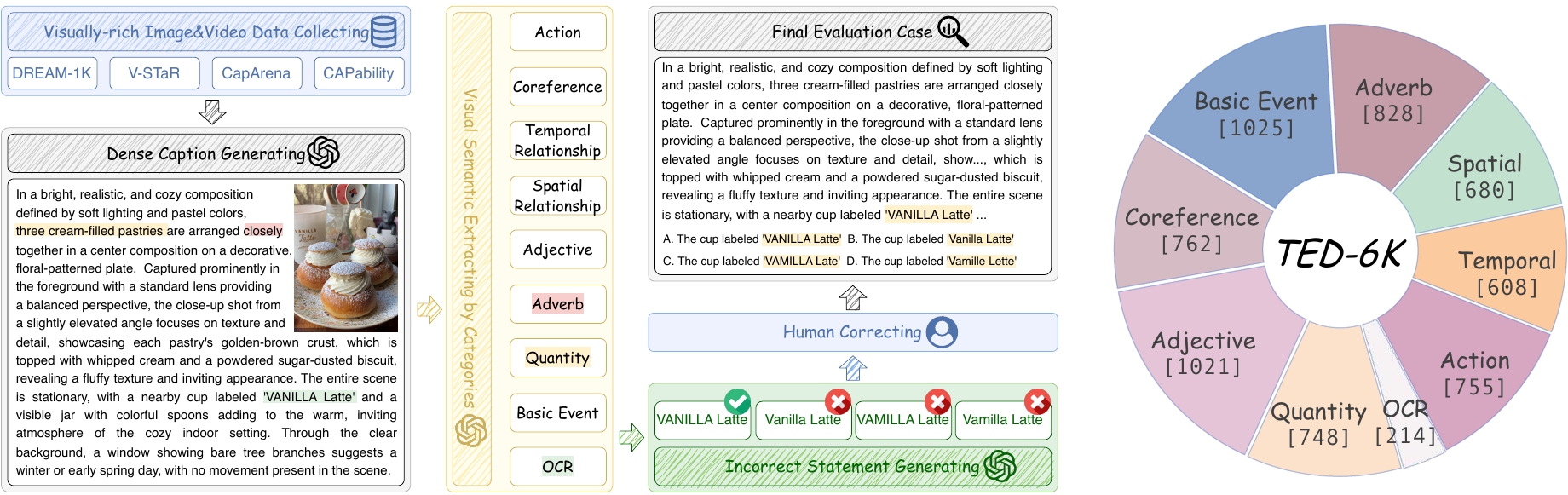}
    \caption{\textit{Left.} The data construction pipeline for TED-6K, consisting of four stages: (1) Data Curation and Filtering; (2) Base Caption Generation; (3) Semantic Pair Construction; (4) Human Verification. \textit{Right.} The data Composition of the TED-6K dataset.}
    \label{fig:pipeline_data}
\end{figure*}


\subsection{Text-only Benchmark for Evaluation}
\label{sec:text-only benchmark}

We first build a text-only benchmark TED-6K illustrated in \cref{fig:pipeline_data}. Our goal is to systematically create a dataset capable of probing a text encoder's representational capabilities across key semantic dimensions. To this end, we design and execute the following multi-stage data pipeline:

\begin{itemize}
\item \textbf{Raw Data Curation and Filtering:} We begin by collecting raw image and video from several high-quality open-source datasets, including DREAM1K~\cite{wang2024tarsier}, CAPability~\cite{liu2025good}, V-STaR~\cite{cheng2025v}, and CapArena~\cite{cheng2025caparena}. These sources are selected for their diversity, high visual fidelity, and semantic complexity.
\item \textbf{Generation of Fine-Grained Base Captions:} For each filtered visual sample, we utilize the advanced capabilities of Gemini 2.5 Pro~\cite{comanici2025gemini} to generate a fine-grained descriptive caption. We guide the model to produce descriptions of the utmost detail, while ensuring maximum coverage of our predefined semantic categories.
\item \textbf{Construction of Multi-Dimensional Semantic Evaluation Pairs:} To probe specific semantic capabilities, we construct evaluation data for each base caption from eight predefined semantic perspectives: action, spatial relationship, temporal relationship, coreference, adjectives, adverbs, quantity, OCR, and basic event. First, if the original base caption contains information that could be articulated from a given perspective, we prompt the model to generate one positive statement consistent with its content. Subsequently, to construct high-quality negative samples, the model performs three distinct types of targeted semantic modifications on this positive statement, such as attribute swapping or relation reversal. A critical constraint is that each modification must create an explicit semantic contradiction with the original base caption, generating plausible yet factually incorrect statements.
\item \textbf{Rigorous Human Verification:} Finally, a manual two-step verification is conducted: annotators first confirm each positive statement matches the original caption, then ensure negative statements are highly confusable with the positive ones. They explicitly target ``hard-negatives'' featuring high lexical similarity and subtle character variations.
\end{itemize}

This pipeline culminates in a dataset of 6,641 evaluation instances, each comprising a source caption, a corresponding positive statement, and a set of semantically contradictory negative statements. The specific distribution of evaluation instances across different semantic categories is illustrated in \cref{fig:pipeline_data}.

\subsection{Sentence-level Context Aggregator}
\label{sec:adapter}

\begin{wrapfigure}{r}{0.5\textwidth} 
    \centering
    \includegraphics[width=0.48\textwidth]{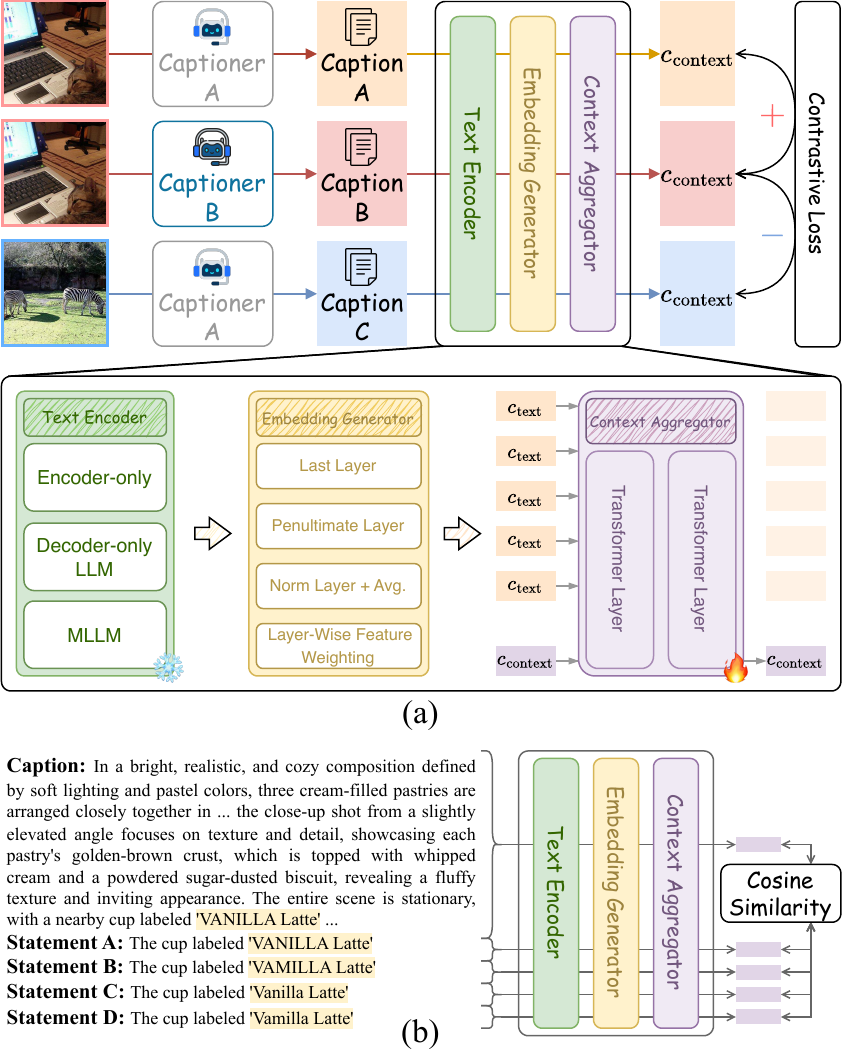} 
    \caption{The context aggregator architecture and its training\&inference process.}
    \label{fig:attn_pool}
\end{wrapfigure}

It is difficult to fairly evaluate the representation capability of different kinds of text encoders with traditional metics, like single-vector retrieval or Question-Answer (QA) accuracy. Most text encoders like LLM or MLLMs generally exhibit suboptimal single-vector retrieval performance; furthermore, this single-vector paradigm inherently misaligns with how contemporary diffusion models utilize text embeddings. Meanwhile, evaluating text encoders for diffusion models via question-answering (QA) accuracy is misleading. This approach is problematic because it cannot be applied to encoders like CLIP that lack QA functionality. More importantly, it conflates two distinct mechanisms within LLMs. QA relies on causal decoding using prefilling features from all model layers to infer an answer, whereas a Diffusion Transformer (DiT) typically uses a static embedding  as its conditioning. Therefore, an LLM's reasoning prowess in QA is not a reliable proxy for its ability to generate high-quality representations for visual generation.

To fairly assess the representation capability of different text encoders when used in DiTs, we design a unified context aggregator that mimics how DiTs consume text embeddings, and evaluate them at the representation level on TED-6K. The context aggregator, $\mathcal{A}_{\text{context}}$, consists of two self-attention Transformer layers and a learnable context token, $c_{\text{context}}$. It functions by aggregating the sentence-level semantics from an input text representation, $c_{\text{text}}$, into the context token. This information flow can be represented as:
\begin{equation}
    c'_{\text{context}} \xleftarrow{} \mathcal{A}_{\text{context}}([c_{\text{text}}, c_{\text{context}}])
\end{equation}
where $c'_{\text{context}}$ is the updated context token, now infused with the information from $c_{\text{text}}$. The detailed architecture is provided in the Appendix A.

~\cref{fig:attn_pool} illustrates the overall workflow of our evaluation framework. To construct the positive pairs for training, we employed the following strategy for each image: using the Qwen3-VL-235A22B-Instruct, we generated two semantically similar but textually diverse captions by prompting it with two distinct prompts at a temperature of 1.0. Overalll, we collect 500k caption pairs from DenseFusion~\cite{li2024densefusion}, ScaleCap~\cite{xing2025scalecap}, MiraData~\cite{MiraData}, Koala-36M~\cite{wang2025koala}, and Video-UFO~\cite{wang2025videoufo}. Notably, the data in the TED-6K benchmark is distinct from our fine-tuning dataset, ensuring a fair evaluation of the model's generalization capabilities.

To enable the context aggregator $\mathcal{A}_{\text{context}}$ to effectively aggregate core sentence-level semantics from the sequence of hidden states, we train it using a contrastive loss~\cite{oord2018representation}. First, both captions from a positive pair are passed through a frozen text encoder (using various feature extraction configurations like Last Layer or Norm-Avg) to obtain their respective textual representations, $c_{\text{text}}^A$ and $c_{\text{text}}^B$. Then, the aggregator, $\mathcal{A}_{\text{context}}$, processes each of these representations to produce their corresponding context tokens, $c'_{\text{context, A}}$ and $c'_{\text{context, B}}$. Finally, we apply the contrastive loss, which is trained to pull together the context tokens from the same image while pushing apart those from different images. 


During the evaluation on TED-6K, a source caption and its candidate statements are passed through a frozen text encoder,  feature extractor and the context aggregator that were specifically trained for it, to obtain their respective context tokens. An evaluation instance is marked as correct if the similarity score between the source caption and the positive statement is the highest among all candidates. The final accuracy is the proportion of correctly identified instances.


\subsection{Specializing MLLMs as Text Encoders for Visual Generation}
\label{sec:our model}

Upon the context aggregator and TED-6K, we find that MLLMs exhibit a significant advantage over other models when used as text encoders for generative tasks, detailed in ~\cref{sec:text-only results}. This superiority is largely attributable to their multimodal training, which forces the text encoder to learn a semantic space that inherently aligns with visual concepts.

However, the training objectives of most existing MLLMs are geared towards developing general-purpose multimodal reasoning capabilities, with their training data often incorporating tasks like visual grounding and multimodal reasoning. These tasks primarily enhance a model's reasoning abilities, while its core function as a text encoder for generative tasks—the capacity to provide high-quality textual representations—is not explicitly optimized.

These analyses raise a natural question: can we develop an even more potent model by fine-tuning a strong MLLM backbone on tasks specifically tailored for visual generation? Motivated by this hypothesis, we introduce GRAN-TED, a text encoder specialized for visual generation, which we developed by building upon the Qwen3-VL-8B-Instruct model. Specifically, we adopt a data-centric approach~\cite{bai2024survey} to enhance the model's capability on text embeddings. We have meticulously collected a large-scale dataset of figures/videos and their corresponding semantically rich captions. Based on this collection, we construct a massive set of Visual Question Answering (VQA) pairs from multiple perspectives directly relevant to visual generation, such as object attributes, spatial relationships, and temporal order. By fine-tuning the Qwen3-VL-8B-Instruct model on this highly targeted VQA dataset, we develop GRAN-TED, a text encoder enhanced for diffusion models.


\subsection{Layer-wise Feature Weighting}
\label{sec:feature combi}

Recent empirical studies~\cite{wang2025comprehensive} demonstrate that normalizing and averaging hidden states (Norm-Avg) significantly outperforms simple averaging (Avg).  In pre-norm LLMs, where the norm of hidden states tends to increase with layer depth~\cite{kim2025peri,li2025unseen},  the Norm-Avg operation not only serves to stabilize the numerical scale of the features but also functions as an implicit, inverse-norm weighting scheme. 

This insight inspires us to hypothesize that if a fixed, implicit weighting is beneficial, an explicit and learnable weighting mechanism could achieve superior performance. To this end, we propose a learnable, layer-wise feature weighting module. The module first applies layer normalization to the hidden state $h_i$ of each layer. It then assigns a set of learnable scalar weights $w_i$ to these normalized representations, which are converted into normalized aggregation weights  $\alpha_i$ via a softmax function:
\begin{equation}
    \alpha_i = \frac{\exp(w_i)}{\sum_{j=1}^{L} \exp(w_j)}
\end{equation}

The final fused representation $c_{\text{text}}$ is then computed as the weighted sum of these normalized representations:
\begin{equation}
   c_{\text{text}} = \sum_{i=1}^{L} \alpha_i \cdot \text{LayerNorm}(h_i) 
\end{equation}
This module allows the model to autonomously learn the importance of features from different hierarchical levels as required by the task, thereby constructing a more informative textual representation.

However, the training objective of a diffusion model evolves: early stages focus on denoising low-frequency global structures, while later stages refine high-frequency details. Given that different layers of an LLM capture distinct types of semantic information~\cite{dar2022analyzing,skean2024does}, the optimal weighting of textual features for these two stages would likely differ. Consequently, allowing the layer weights to update continuously would cause them to drift with the evolving training objective, creating a non-stationary text condition that poses a significant challenge to the stable convergence of the diffusion model (e.g., DiT).

To mitigate this instability, we introduce a two-step training strategy. We first train the layer weights jointly with the diffusion model for an initial number of steps, allowing them to converge to a generally effective scheme. After that, we freeze these learned weights, thereby providing a stable, high-quality text representation for the remainder of the model's training. We provide detailed discussion on the two-step training strategy in the Appendix B.

\section{Experiments}

We evaluate our TED-6K benchmark across several mainstream text encoders. To validate the benchmark's effectiveness—whether its scores can predict real-world performance in downstream generative tasks—we select a representative subset of these encoders and train corresponding T2I and T2V models for a formal correlation analysis. Both the TED-6K and T2I\&T2V benchmarks shows that GRAN-TED has a superior text representation capability. The detailed experimental setup can be seen in Appendix D.

\subsection{TED-6K Results}

\begin{table*}[!htbp]
\centering
\caption{Performance comparison of different text encoders and feature extraction strategies on our text-only benchmark.}
\label{tab:text-only-results}
\resizebox{0.9\linewidth}{!}{%
\begin{tabular}{@{}llccc@{}}
\toprule
\textbf{Model Type} & \textbf{Model Name} & \textbf{Last Layer } & \textbf{Penultimate Layer } & \textbf{Norm-Avg} \\
\midrule
\multirow{2}{*}{\textit{Encoder-only}} & UMT5-XXL~\cite{chung2023unimax} & 44.60 & 46.89 & 51.41 \\
& AltCLIP~\cite{chen2023altclip} & 47.15 & 46.59 & 51.86 \\
\midrule
\multirow{7}{*}{\textit{LLM}} & Qwen3-8B-Instruct & 53.62 & 53.00 & 55.77 \\
& Qwen3-8B-Base & 53.37 & 54.19 & 55.94 \\
& Qwen3-4B-Instruct &  53.64 & 53.58 & 55.25  \\
& Qwen3-4B-Base &  52.84 & 53.06 & 54.98  \\
& Qwen3-4B-Thinking & 52.90 & 54.10 & 54.86 \\
& Qwen3-32B-Instruct & 53.23 & 54.42 & 56.98 \\
& Llama3-8B-Instruct~\cite{grattafiori2024llama} & 54.75 & 55.22 & 55.93 \\ 
& InternLM3-8B-Instruct~\cite{cai2024internlm2} & 55.10 & 55.40 & 55.46 \\
\midrule 
\multirow{8}{*}{\textit{MLLM}} & Qwen3-VL-8B-Instruct & 55.37 & 54.25 & 56.81 \\
& Qwen3-VL-8B-Thinking & 55.67 & 55.40 & 56.51 \\
& Qwen3-VL-4B-Instruct & 54.03 & 54.92 & 55.20  \\
& Qwen3-VL-32B-Instruct & 54.60 & 55.91 & 57.24 \\
& MiMo-VL-7B~\cite{xiaomi2025mimo} & 54.54 & 56.29 & 56.48 \\
& Ovis-2.5-9B~\cite{lu2025ovis2} & 54.66 & 54.92 & 56.26 \\
& Gemma3-4b-it~\cite{kamath2025gemma} & 54.52 & 53.80 & 54.59 \\
& InternVL3-9B-Instruct~\cite{zhu2025internvl3} & 55.68 & 55.68 & 55.59\\
\midrule
\rowcolor{gray!20} \multirow{1}{*}{} & GRAN-TED w/o weighting & 55.87 & 55.13 & 57.22 \\
\rowcolor{gray!20} \multirow{1}{*}{} & GRAN-TED & - & - & \textbf{57.42} \\
\bottomrule
\end{tabular}
}
\end{table*}

\label{sec:text-only results}
 \cref{tab:text-only-results} presents the evaluation results of various text encoder types on our text-only benchmark. Fine-tuned on our curated dataset, GRAN-TED not only outperforms all peer models but also remarkably closes the gap with 32B Model.

A detailed analysis of  \cref{tab:text-only-results}  reveals several clear and insightful trends:
\begin{itemize}
    \item \textbf{Potential of Decoder-only Architectures:} The benchmark results on TED-6K unequivocally reveal that decoder-only LLMs exhibit significantly superior representational capabilities as text encoders compared to traditional encoder-only models. Notably, this performance gap does not stem solely from model size. For instance, the 5B-parameter UMT5-XXL encoder still performs considerably worse than the smaller 4B Qwen3-VL-Instruct.
    \item \textbf{Value of Multimodal Training:} A key and noteworthy finding is that MLLMs consistently and significantly outperform their LLM backbones on our text-only benchmark, despite the evaluation itself being entirely devoid of visual inputs. We believe this advantage stems from the multimodal training process, which compels the text encoder to learn how to more effectively encode visually-grounded concepts within its hidden states, thereby enabling the language model component to better leverage these representations for downstream tasks like VQA.
    
    \item \textbf{Ambiguous Effect of Instruction Tuning:}  In contrast to the clear trends observed in multimodal settings, instruction tuning does not exhibit a consistent impact on the representational capabilities  on TED-6K.
    
    \item \textbf{Impact of Thinking model} Interestingly, thinking models exhibit a paradoxical effect: while they enhance the representational quality of single-layer features, they conversely degrade performance when multi-layer features are aggregated via the Norm-Avg strategy. This likely stems from their distinct reasoning behavior during QA, which fundamentally alters their prefill hidden states and the resulting representations.
    \item \textbf{Effective Scaling Trends:} The scaling law on the model size differs significantly depending on the feature extraction strategy. When relying solely on single-layer features, we observed no clear scaling trend. However, when using the Norm-Avg strategy, the model's score on TED-6K exhibits a much clearer positive correlation with its parameter size.
    \item \textbf{Importance of Feature Aggregation:} Comparing single-layer features, the penultimate layer lacks a consistent advantage over the final layer. However, aggregating multi-layer features via the Norm-Avg strategy generally outperforms using any single layer on TED-6K. Since different layers capture distinct syntactic and semantic granularities, their aggregation typically yields a richer text representation and superior overall performance.
    
\end{itemize}

\begin{table*}[t]
\centering
\caption{Performance breakdown of Qwen3-VL and GRAN-TED across nine semantic sub-tasks (Norm-Avg).}
\label{tab:ted6k_breakdown}
\small 
\setlength{\tabcolsep}{0pt}
\begin{tabular*}{\textwidth}{@{\extracolsep{\fill}}lccccccccc@{}}
\toprule
\textbf{Model} & \textbf{Quant.} & \textbf{Adj.} & \textbf{Coref.} & \textbf{Event} & \textbf{Adv.} & \textbf{Spat.} & \textbf{OCR} & \textbf{Temp.} & \textbf{Act.} \\
\midrule
Qwen3-VL & 52.01 & 46.91 & 55.77 & 76.98 & 47.71 & 46.32 & 35.98 & 63.32 & 68.87 \\
GRAN-TED & 52.87 & 45.74 & 57.16 & 77.86 & 45.73 & 47.11 & 37.12 & 65.99 & 71.19 \\
\cmidrule(lr){1-10} 
$\Delta$ & +0.86 & -1.17 & +1.39 & +0.88 & -1.98 & +0.79 & +1.14 & +2.67 & +2.32 \\
\bottomrule
\end{tabular*}
\end{table*}

 \cref{tab:ted6k_breakdown} presents a detailed performance breakdown of the models on the various sub-tasks of TED-6K. GRAN-TED outperforms the baseline model across all semantic dimensions except for Adjective and Adverb. Further analysis indicates that the model excels at capturing macro-level events (e.g., ``Action'') but still has room for improvement in dimensions requiring fine-grained detail (e.g., ``Spatial Relationship'', ``OCR''), highlighting a clear direction for future work.

\subsection{Correlation with Generation Performance}
\label{sec:correlation}

To validate our text-only benchmark and investigate the consistency of its scores with downstream generative performance, we conduct a quantitative analysis by calculating the Pearson correlation coefficient between our benchmark scores and downstream evaluations. For these downstream metrics, we utilize GenAI-Bench~\cite{li2024genai} for both T2I and T2V tasks, alongside UnifiedReward-2~\cite{unifiedreward} scores on DrawBench~\cite{Saharia2022PhotorealisticTD} for T2I, and VideoAlign~\cite{liu2025improving} rewards on TA-hard\cite{liu2025improving} for T2V. The comprehensive correlation results for the T2I and T2V tasks are presented in  \cref{tab:T2I_correlation} and  \cref{tab:T2V_correlation}, respectively. We observe that the scores from our text-only benchmark exhibit a statistically significant and strong positive correlation with the models' generative performance across all datasets. Furthermore, we include examples of images generated on GenAI-Bench using these models as text encoders in the Appendix C.

\begin{table}[!htbp]
\centering

\caption{Correlation between our benchmark scores and T2I generation performance. All evaluated encoders are instruction models, except for UMT5-XXL.} 
\label{tab:T2I_correlation}

\begin{tabular}{@{}lccc@{}}
\toprule
\textbf{Text Encoder} & \textbf{TED-6K} & \textbf{GenAI}$\uparrow$ & \textbf{UnifiedReward}$\uparrow$ \\
\midrule
UMT5-XXL {\small (last-layer)}           & 44.60 & 54.67 &  2.834 \\
Qwen2.5-VL-7B {\small (last-layer)}      & 53.05 & 71.15 & 2.995 \\
Qwen3-8B {\small (last-layer)}           & 53.62 & 72.42 & 2.994 \\
Gemma3-4b-it {\small (last-layer)}       &   54.52   &  71.56   & 2.999 \\
InternVL3.5-8B-MPO {\small (last-layer)} & 55.31 & 74.46 & 3.003 \\
Qwen2.5-VL-7B {\small (norm-avg)}        & 55.99 & 75.51 & 2.997 \\
Qwen3-VL-8B {\small (norm-avg)}          & 56.81 & 76.17 & 3.028 \\
\textbf{GRAN-TED} & \textbf{57.42} & \textbf{77.41} & \textbf{3.035}\\
\midrule
\rowcolor{gray!20} \multicolumn{2}{l}{\textit{Pearson Corr. ($r$) w/ TED-6K}} & 0.9923 & 0.9778 \\
\rowcolor{gray!20} \multicolumn{2}{l}{\textit{$p$-value}} & $1.11\times 10^{-6}$ & $2.69 \times 10^{-5}$ \\
\bottomrule
\end{tabular}

\end{table}
\begin{table}[!htbp]
\centering
\caption{Correlation between our benchmark scores and T2V generation performance. } 
\label{tab:T2V_correlation}
\begin{tabular}{@{}lccc@{}}
\toprule
\textbf{Text Encoder} & \textbf{TED-6K} & \textbf{GenAI}$\uparrow$ & \textbf{VideoAlign}$\uparrow$ \\
\midrule
Qwen3-8B {\small (last-layer)}           & 53.62 & 65.13 & 0.767 \\
Qwen3-VL-8B {\small (last-layer)}        & 55.37 & 70.59 & 0.967 \\
InternVL3.5-8B-MPO {\small (last-layer)} & 55.31 & 68.70 & 1.002 \\
Qwen3-VL-8B {\small (norm-avg)}          & 56.81 & 77.94 & 1.288 \\
\textbf{GRAN-TED} & \textbf{57.42} & \textbf{80.33} & \textbf{1.424}\\
\midrule
\rowcolor{gray!20} \multicolumn{2}{l}{\textit{Pearson Corr. ($r$) w/ TED-6K}} & 0.9739 & 0.9866 \\
\rowcolor{gray!20} \multicolumn{2}{l}{\textit{$p$-value}} & $5.05\times 10^{-3}$ & $1.87\times 10^{-3}$ \\
\bottomrule
\end{tabular}
\end{table}

\subsection{Ablation Study}
To validate the effectiveness of the proposed method, we further select Qwen3-VL-8B-Instruct as the base model for our experiments. The results can be seen in  \cref{tab:two-stage-ablation}.
As shown in the table, our two-step training strategy proposed in  ~\cref{sec:feature combi} significantly improves performance over the vanilla Norm-Avg, yet it only introduces a small number of learnable parameters equal to the number of Transformer layers in the LLM. 

\begin{wrapfigure}{r}{0.5\textwidth} 
    \centering
    \includegraphics[width=0.48\textwidth]{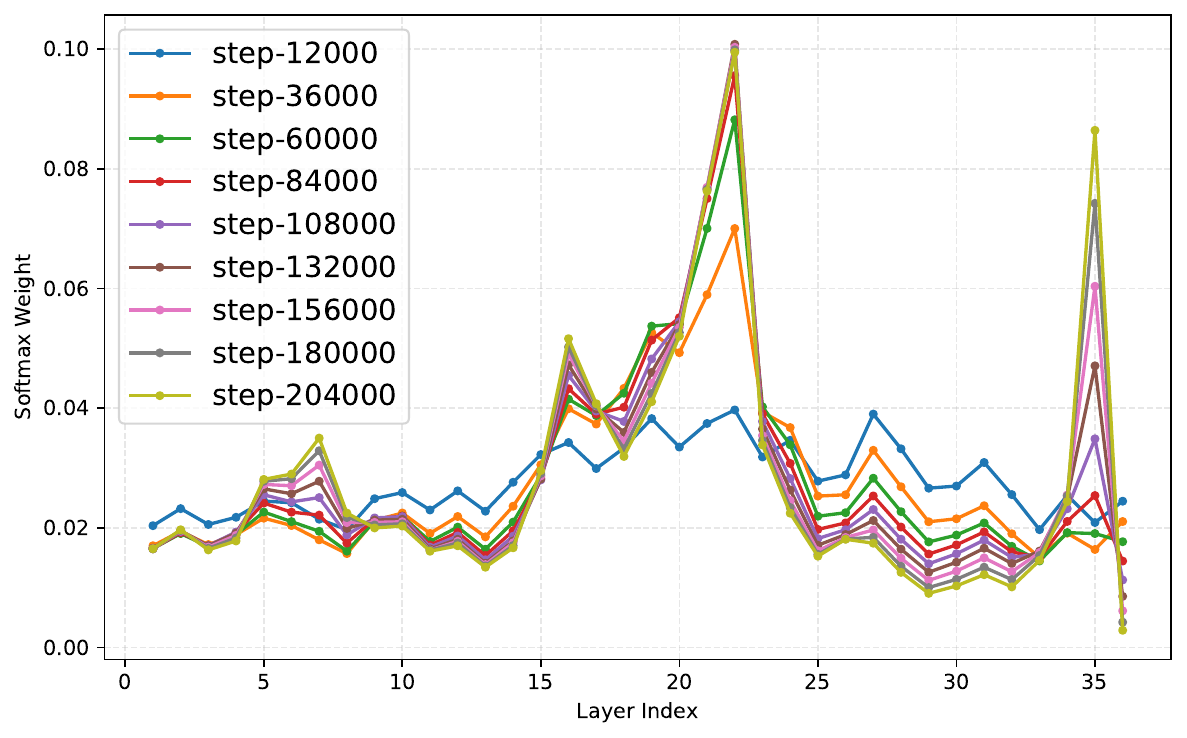}
    \caption{Dynamics of the learnable layer weights over the course of continuous training (i.e., without the two-step strategy). The weight values shown are normalized via Softmax.}
    \label{fig:weight_dynamics}
\end{wrapfigure}

Notably, when the two-step strategy is omitted and only learnable weights are introduced with continuous training, performance slightly degrades.
This observation aligns with the argument in  \cref{sec:feature combi} that a non-stationary text condition can interfere with the training process.
We visualize the weights assigned to each layer at different training steps during the training progress.
As illustrated in  \cref{fig:weight_dynamics}, the evolution of the layer weights reveals two key phenomena.

First, the weights of different layers exhibit non-monotonic dynamics. For instance, in the shallower regions of the model (e.g., layers 5 through 8), the weights for these layers collectively first increase and then decrease over the course of training. 
Second, the overall weight distribution displays persistent instability, failing to converge to an optimal solution. Even after 200k training steps, some deeper layers, such as the penultimate layer (-2), are still undergoing a substantial and consistent increase, while the weights of many other layers remain in flux. This observation provides a direct visual confirmation of our hypothesis that the text condition is indeed non-stationary under a continuous training paradigm, thereby underscoring the necessity of our second-stage weight-freezing strategy.

\begin{table}[ht]
\centering
\caption{Ablation study over Qwen3-VL-8B-Instruct model. ``Learnable Weights": trainable layer weights with continuous training. ``Two-Step Training": the two-step training strategy to the learnable weights.}
\label{tab:two-stage-ablation}
\begin{tabular}{@{}lcc@{}}
\toprule
\textbf{Method} & \textbf{GenAI-Bench} & \textbf{UnifiedReward} \\
\midrule
Norm-Avg (Fixed Weights) & 76.17 & 3.028 \\
+ Learnable Weights & 75.94 & 3.021 \\
+ Two-Step Training  & \textbf{77.01} & \textbf{3.035} \\
\bottomrule
\end{tabular}
\end{table}

Finally, we train diffusion models integrated with GRAN-TED, enhanced by both Norm-Avg and two-step layer-wise feature weighting. 
We then compare these models against a strong baseline trained on the Norm-Avg representations from the original Qwen3-VL-8B. 
The results demonstrate that models conditioned on our full methodology achieve superior generation performance across both modalities. 
For the T2I task (Table~\ref{tab:T2I_correlation}), GRAN-TED improves upon the baseline by +1.24 in GenAI and +0.007 in UnifiedReward. 
Similarly, for the T2V task (Table~\ref{tab:T2V_correlation}), it achieves even more significant gains of +2.39 in GenAI and +0.136 in NewBench. 
These downstream improvements are consistent with the +0.61 absolute increase in our TED-6K benchmark, underscoring the robust representational alignment capabilities of GRAN-TED.


\subsection{Further Analysis}

\noindent\textbf{Performance Discrimination from TED-6K.}
The models achieve relatively similar scores as shown in ~\cref{tab:text-only-results}. The reason is that part of the caption-statement pairs are simple for a comprehensive assessment, resulting in a low representational challenge. However, on the more demanding downstream tasks (~\cref{tab:T2I_correlation}-\cref{tab:T2V_correlation}), which concentrate on difficult prompts, the performance gaps between models become significantly more pronounced. This validates that TED-6K is effectively discriminative.

\noindent\textbf{Robustness of TED-6K.}
To ascertain the contextual dependency of the questions in TED-6K, we decouple the captions from their corresponding statements and make shuffle. The Qwen3VL-8B-Instruct model is then tasked with direct question-answering on this shuffled data. Across five experimental runs, the model achieves an accuracy between 27.05\% and 28.63\%. This performance is approximately at the level of random guessing (25\% for a four-option task), which demonstrates that the questions in TED-6K are sufficiently challenging and require the correct contextual caption to be answered accurately.

\noindent\textbf{Aggregator v.s. Single-vector Representation.}
A comparison between using a trained aggregator and applying mean pooling over the final layer's outputs reveals divergent results across architectures. According to ~\cref{tab:furtheranalysis}, while mean pooling improves representation quality for Encoder-only model, it still falls short of the best performance on TED-6K and exhibits a trend inconsistent with downstream tasks. Conversely, for Decoder-only LLMs, this method markedly impairs representational power. This justifies the necessity of our approach, which involves training an aggregator for effective model validation.

\noindent\textbf{Defectiveness of QA-style Assessment.}
We compare the proposed similarity-based method with a direct QA method. For the QA task, we prompt the Qwen2.5-72B model to first convert each sample's statements into a question and then ask the text encoders to provide an open-ended answer. As detailed in ~\cref{tab:furtheranalysis}, the model's powerful reasoning lead to uniformly accurate answers, which diminishes the method's ability to distinguish between test cases. This demonstrates the suitability of the similarity-based approach as a more discriminative evaluation metric.

\noindent\textbf{Stability of TED-6K.}
To account for the variability from model training in our Aggregator approach, we train the aggregators for UMT5-XXL, Qwen3-8B-Instruct, and Qwen3VL-8B-Instruct five times each. The maximum score variation across these runs is only 0.02 for any given model. This level of variance would not alter the assessed document order relations in the TED-6K benchmark, which confirms the stability of our evaluation methodology.

\begin{table}[ht]
\centering
\caption{Performance comparison of our aggregator-based method against mean-pooling approach and QA-style evaluations.}
\label{tab:furtheranalysis}
\begin{tabular}{@{}lccc@{}}
\toprule
\textbf{Model} & \textbf{ Aggregator} & \textbf{Mean-pooling} & \textbf{Q\& A}\\
\midrule
UMT5-XXL & 44.65 & 50.64  & 97.81 \\
Qwen3-8B-Instruct & 53.62 & 39.18 & 97.90 \\
Qwen3-VL-8B-Instruct & 55.37 & 42.59 & 97.94 \\
\bottomrule
\end{tabular}

\end{table}

\section{Conclusion}
In this work, we address the critical role of the text encoder in T2I and T2V diffusion models. First, we establish an efficient evaluation framework centered on our novel text-only benchmark, TED-6K, for rapid encoder assessment. Second, we propose a two-stage paradigm to develop the GRAN-TED encoder, featuring targeted visual-semantic finetuning and layer-wise weighting. Experiments validate that GRAN-TED achieves state-of-the-art performance on TED-6K, translating into demonstrable improvements in semantic accuracy and compositional coherence for both T2I and T2V generation.



%
%
\bibliographystyle{splncs04}
\bibliography{main}

\clearpage

\appendix
\renewcommand{\theHsection}{appendix.\Alph{section}}
\renewcommand{\theHsubsection}{appendix.\Alph{section}.\arabic{subsection}}
\renewcommand{\theHfigure}{appendix.\thefigure}
\renewcommand{\theHtable}{appendix.\thetable}
\section{Aggregator Architecture}
\label{sec:aggretator_arch}

We implement the sentence-level context aggregator $\mathcal{A}_{\text{context}}$ as a lightweight attention-based pooling module. Given token-level hidden states from a frozen text encoder, $\mathcal{A}_{\text{context}}$ produces a single sentence-level context embedding $c'_{\text{context}}$ that mimics the text conditioning consumed by a DiT.

\paragraph{Inputs.}
Let $c_{\text{text}} \in \mathbb{R}^{B \times N \times D}$ denote a sequence of $N$ token embeddings of dimension $D$ for a batch of size $B$. The aggregator supports two kinds of inputs:
(i) single-layer features $c_{\text{text}} \in \mathbb{R}^{B \times N \times D}$ and
(ii) stacked multi-layer features $\{h_i\}_{i=1}^{L}$ with $h_i \in \mathbb{R}^{B \times N \times D}$ from $L$ encoder layers.

\paragraph{Layer-wise fusion.}
When multi-layer representations are available, $\mathcal{A}_{\text{context}}$ first fuses them into a single token sequence before applying attention. Each layer output is normalized with a parameter-free LayerNorm, yielding $\{\tilde{h}_i\}_{i=1}^{L}$, and the normalized features are either
(i) averaged across layers, or
(ii) combined using a learnable softmax-weighted sum. Concretely,
\begin{equation}
    c_{\text{text}}
    =
    \begin{cases}
        \dfrac{1}{L}\sum_{i=1}^{L} \tilde{h}_i, & \text{(average)},\\[4pt]
        \sum_{i=1}^{L} \alpha_i \tilde{h}_i,\quad
        \alpha_i = \dfrac{\exp(\omega_i)}{\sum_{j=1}^{L} \exp(\omega_j)}, & \text{(learnable)},
    \end{cases}
\end{equation}
where $\omega = \{\omega_i\}_{i=1}^{L}$ are scalar layer weights (optionally trainable). For single-layer inputs this fusion step is skipped and $c_{\text{text}}$ is simply the encoder hidden states.

\paragraph{context token and attention stack.}
Given the fused sequence $c_{\text{text}} \in \mathbb{R}^{B \times N \times D}$, $\mathcal{A}_{\text{context}}$ prepends a learnable context token $c_{\text{context}} \in \mathbb{R}^{1 \times 1 \times D}$ (shared across the batch):
\begin{equation}
    x_0 = \mathrm{concat}(c_{\text{context}},\, c_{\text{text}})
    \in \mathbb{R}^{B \times (1+N) \times D}.
\end{equation}
This context token serves as a global context carrier. The resulting sequence is processed by a stack of $L_{\text{blk}}$ pre-LayerNorm Transformer blocks, each consisting of multi-head self-attention followed by a two-layer MLP with GELU activation and residual connections. A padding mask (when available) is applied in a bidirectional way: padded positions are ignored in attention, while all valid tokens and the context token can freely attend to each other. This two-block self-attention stack implements the core of the aggregator architecture.

\paragraph{Output.}
After the attention stack, we obtain $x_{\text{enc}} \in \mathbb{R}^{B \times (1+N) \times D}$. We apply a final LayerNorm and, if needed, a linear projection to match the target dimension $D_{\text{out}}$:
\begin{equation}
    x_{\text{proj}}
    = \mathrm{Proj}(\mathrm{LayerNorm}(x_{\text{enc}}))
    \in \mathbb{R}^{B \times (1+N) \times D_{\text{out}}}.
\end{equation}
The aggregated context embedding is read out from the projected context token:
\begin{equation}
    c'_{\text{context}} = x_{\text{proj}}[:, 0, :] \in \mathbb{R}^{B \times D_{\text{out}}}.
\end{equation}
In summary, the aggregator instantiates the abstract mapping
\begin{equation}
    c'_{\text{context}}
    \leftarrow \mathcal{A}_{\text{context}}\bigl([c_{\text{text}},\, c_{\text{context}}]\bigr),
\end{equation}
where $c_{\text{text}}$ denotes token-level text embeddings and $c_{\text{context}}$ is represented by the learnable context token.

\section{Theoretical Justification for the Two-Step Training Strategy}
\label{sec:two-stage-argument}

We present a formal argument for our two-step training strategy, grounded in the principles of optimization dynamics and the spectral bias of neural networks. The argument proceeds in a clear logical chain: we first establish the asymmetric roles of the model parameters $\theta$ and the layer weights $\omega$. We then introduce spectral bias as the mechanism driving an evolving optimization objective. This leads to the deduction of a perpetually moving target for $\omega$, which results in a non-stationary conditioning distribution. Finally, we present the two-step strategy as a practical resolution to this optimization pathology.

\subsection{Premise 1: Asymmetric Roles in Joint Optimization}

Let $\theta$ be the parameters of the denoising network $\epsilon_\theta$, and $\omega$ be the learnable layer weights of our text fusion module $f_\omega$ that produces the context vector $c_{\text{context}}$. While jointly optimized, their roles are fundamentally asymmetric. The goal is to minimize the loss $\mathcal{L}(\theta, \omega)$:
\begin{equation}
    \label{eq:joint_loss_appendix}
    \begin{multlined}
    \min_{\theta, \omega} \mathcal{L}(\theta, \omega) \\
    \hfill
    = \min_{\theta, \omega} \mathbb{E}_{c_{\text{context}} \sim P_{c_{\text{context}}}(\cdot; \omega)}
    \Bigl[
        \mathbb{E}_{x_0, \epsilon, t}
        \bigl\|
            \epsilon - \epsilon_\theta(x_t, t, c_{\text{context}})
        \bigr\|^2
    \Bigr], \\
    \hfill
    \text{where } c_{\text{context}} = f_\omega(\text{text embeddings}).
    \end{multlined}
\end{equation}
Here, $P_{c_{\text{context}}}(\cdot; \omega)$ is the distribution of the conditioning vector $c_{\text{context}}$ produced by $f_\omega$.

\begin{itemize}
    \item \textbf{The role of $\theta$:}
    The parameters $\theta$ define a \emph{function approximator}. Their optimization answers the question:
    “Given a fixed input distribution $P_{c_{\text{context}}}(\cdot; \omega)$, how should the function
    $\epsilon_\theta$ change to best predict the noise?”
    \item \textbf{The role of $\omega$:}
    The parameters $\omega = \{\omega_i\}$ are \emph{distributional control parameters}. Their optimization answers the question:
    “For the current function $\epsilon_\theta$, how should the input distribution
    $P_{c_{\text{context}}}$ be shaped (via the layer-wise weights $\omega_i$ in $f_\omega$)
    to minimize the overall loss?”
\end{itemize}

This asymmetry is the foundational premise of our argument.

\subsection{Premise 2: Evolving Objective via Spectral Bias}

The training objective for $\theta$ is not static due to the well-established principle of \textbf{spectral bias} in deep neural networks: networks learn low-frequency functions more rapidly than high-frequency ones.

In the context of diffusion models, this manifests as a natural curriculum:
\begin{enumerate}
    \item \textbf{Initial training:}
    The network $\epsilon_\theta$ first learns to denoise low-frequency components of the signal. This corresponds to capturing the coarse, global structure of the image, which is the dominant source of error at large timesteps $t$ (the low-SNR regime).
    \item \textbf{Later training:}
    Once low-frequency components are learned, the residual error is concentrated in the high-frequency components. The optimization then shifts focus to learning these, which corresponds to refining local details and textures, the primary task at small timesteps $t$ (the high-SNR regime).
\end{enumerate}

Therefore, the denoising task itself evolves, progressing from a low-frequency to a high-frequency focus. This implies that the \emph{type of conditioning information} required by $\theta$ also evolves continuously throughout training. In particular, the effective configuration of the layer weights $\omega_i$ in $f_\omega$ that produce $c_{\text{context}}$ must adapt over time.

\subsection{Deduction: A Continuously Drifting Optimum for the Layer Weights}

Combining our two premises leads to a critical deduction. For any fixed state of the model $\theta_k$, there exists an \textbf{instantaneous optimal configuration} $\omega_k^*$ that shapes the most effective conditioning distribution for that specific state:
\begin{equation}
    \omega_k^* = \arg\min_{\omega} \mathcal{L}(\theta_k, \omega).
    \label{eq:instant_optimum}
\end{equation}

From Premise~2, we know that the task for $\theta_k$ is continuously evolving due to spectral bias. From Premise~1, we know that the role of $\omega$ is to adapt the input distribution to best suit the current $\theta_k$. It logically follows that the instantaneous optimum $\omega_k^*$ must also be in a state of perpetual drift as $\theta_k$ learns:
\begin{equation}
    \omega_k^* \neq \omega_{k+1}^*.
    \label{eq:drifting_optimum}
\end{equation}

This creates a “moving target” problem, where the layer weights $\omega_k$ are forced to perpetually chase an optimum that never settles.

\subsection{Consequence: Non-Stationary Distribution and Instability}

The perpetual chase of a drifting optimum ensures that the layer weights remain in constant flux ($\omega_{k+1} \neq \omega_k$). This directly implies that the conditioning distribution is non-stationary:
\begin{equation}
    \omega_{k+1} \neq \omega_k
    \;\Longrightarrow\;
    P_{c_{\text{context}}}(\cdot; \omega_{k+1})
    \neq
    P_{c_{\text{context}}}(\cdot; \omega_k).
    \label{eq:non_stationary}
\end{equation}

This non-stationarity is a form of \textbf{covariate shift} internal to the training loop, a known cause of optimization pathology.
The gradient $\nabla_\theta \mathcal{L}$ is computed based on a “stale” distribution
$P_{c_{\text{context}}}(\cdot; \omega_k)$ that is immediately invalidated by the update to $\omega_{k+1}$.
This mismatch between the environment used for gradient calculation and the one the updated model faces
leads to inefficient and potentially unstable training.

\subsection{Resolution: The Two-Step Strategy}

The two-step strategy is designed to \textbf{arrest the continuous drift} of the conditioning distribution.

\begin{enumerate}
    \item \textbf{Step 1 (joint optimization).}
    We perform joint optimization over $(\theta, \omega)$ to find a robust $\omega^*$:
    it represents an effective compromise across the continuum of objectives encountered during the initial, dynamic phase of training, as the spectral bias drives $\theta$ from low-frequency to high-frequency behavior.
    \item \textbf{Step 2 (freezing the layer weights).}
    We then \textbf{freeze the layer weights} at this compromise point, i.e., set $\omega = \omega^*$ and keep it fixed for the remainder of training. This halts the evolution of $\omega$ and forces the conditioning distribution to become \textbf{stationary}:
    \begin{equation}
        P_{c_{\text{context}}}(\cdot; \omega_{k+1})
        =
        P_{c_{\text{context}}}(\cdot; \omega_k)
        =
        P_{c_{\text{context}}}(\cdot; \omega^*).
        \label{eq:stationary_dist}
    \end{equation}
\end{enumerate}

By stabilizing the conditioning distribution $P_{c_{\text{context}}}$, we eliminate the internal covariate shift.
The optimization problem for $\theta$ is transformed into a well-posed task on a fixed loss landscape, which facilitates more reliable and efficient convergence.
In practice, this matches the empirical observation in the main paper:
once the layer-wise fusion weights $\omega_i$ have converged to a stable pattern,
further updating them brings marginal benefit while introducing additional noise into the training dynamics of $\theta$.

\section{Image Examples}
\label{app:img}

\newcommand{\StrategyRowSeven}[8]{%
  \setlength{\tabcolsep}{1pt}%
  \renewcommand{\arraystretch}{1}%
  \begin{tabular}{cccccccc}
    {\scriptsize \shortstack{Umt5xxl\\Last Layer}} &
    {\scriptsize \shortstack{Qwen2.5-VL\\Last Layer}} &
    {\scriptsize \shortstack{Qwen3\\Last Layer}} &
    {\scriptsize \shortstack{Gemma3\\Last Layer}} &
    {\scriptsize \shortstack{InternVL3.5\\Last Layer}} &
    {\scriptsize \shortstack{Qwen2.5-VL\\Norm Avg}} &
    {\scriptsize \shortstack{Qwen3-VL\\Norm Avg}} &
    {\scriptsize GRAN-TED} \\
    \includegraphics[width=0.12\linewidth]{#1} &
    \includegraphics[width=0.12\linewidth]{#2} &
    \includegraphics[width=0.12\linewidth]{#3} &
    \includegraphics[width=0.12\linewidth]{#4} &
    \includegraphics[width=0.12\linewidth]{#5} &
    \includegraphics[width=0.12\linewidth]{#6} &
    \includegraphics[width=0.12\linewidth]{#7} &
    \includegraphics[width=0.12\linewidth]{#8} \\
  \end{tabular}%
}
\newcommand{\PromptPanelSeven}[9]{%
  \StrategyRowSeven{#2}{#3}{#4}{#5}{#6}{#7}{#8}{#9}\par%
  {\small Prompt: ``#1''}\par%
  \mbox{}\\[0.5cm]
}

The images below show qualitative comparisons across different text encoders under multiple prompts from GenAIBench~\cite{li2024genai}, all using identical sampling settings to ensure a fair comparison.

\begin{center}

\PromptPanelSeven
  {There are two red flowers on the table and a few yellow flowers under the table.}
  {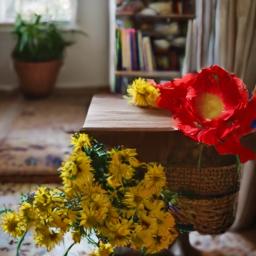}
  {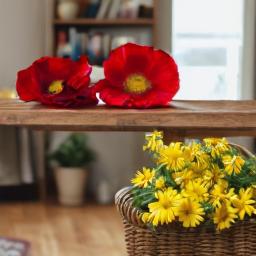}
  {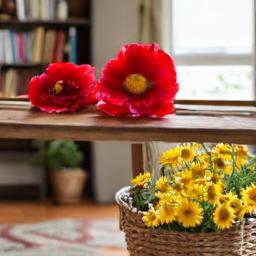}
  {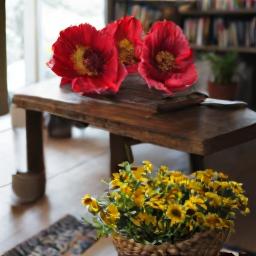}
  {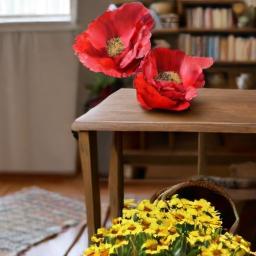}
  {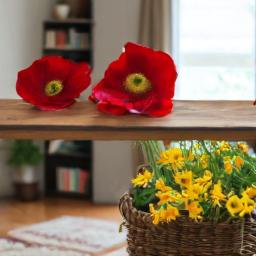}
  {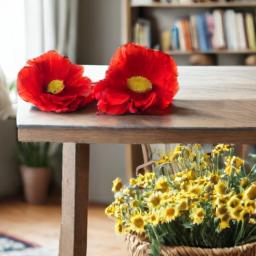}
  {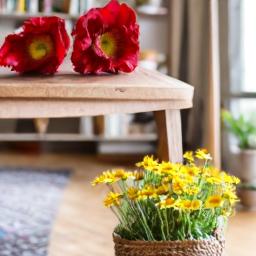}

\PromptPanelSeven
  {Four flowers are in full bloom, with two of them by the bench.}
  {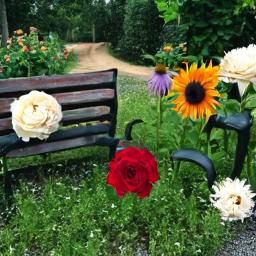}
  {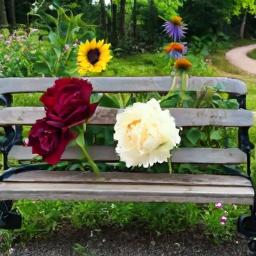}
  {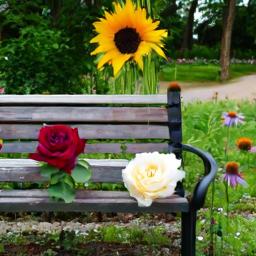}
  {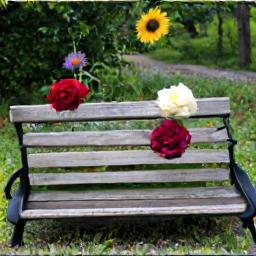}
  {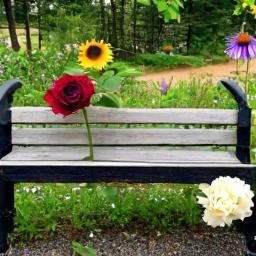}
  {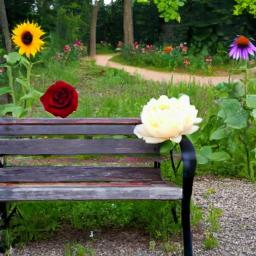}
  {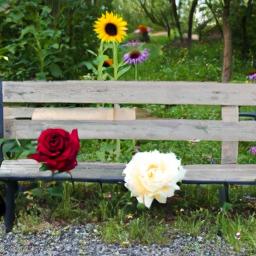}
  {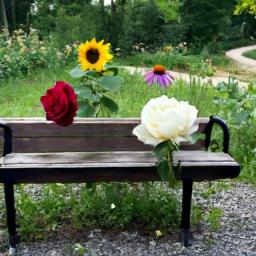}

\PromptPanelSeven
  {Two red balloons and three white clouds floating in the blue sky.}
  {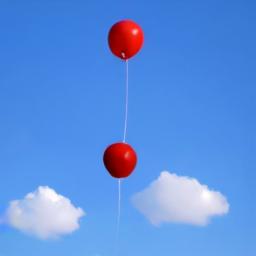}
  {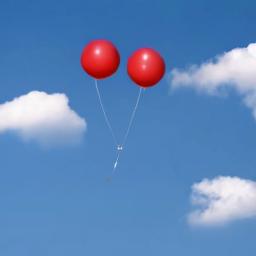}
  {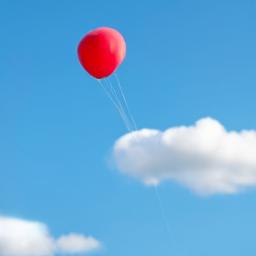}
  {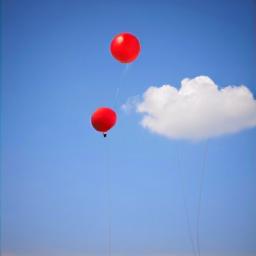}
  {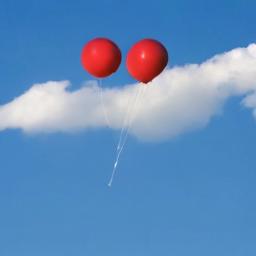}
  {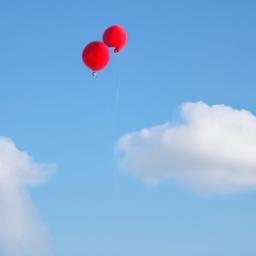}
  {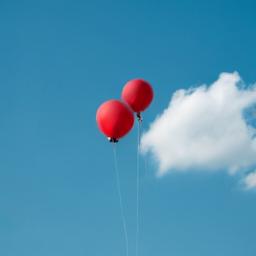}
  {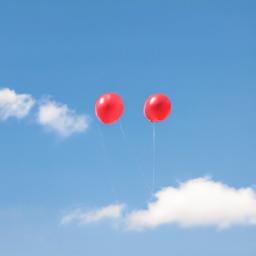}

\PromptPanelSeven
  {Among a group of pastel-colored balloons, one stands out in vibrant red.}
  {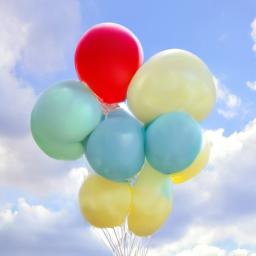}
  {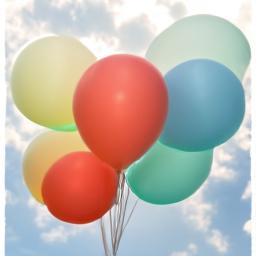}
  {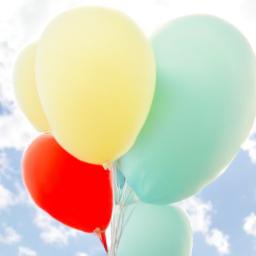}
  {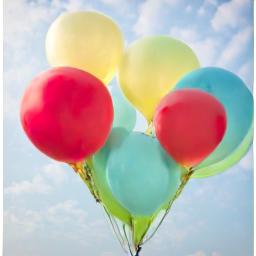}
  {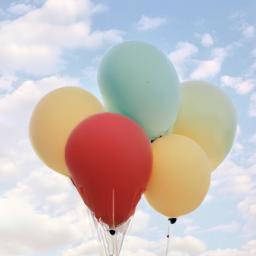}
  {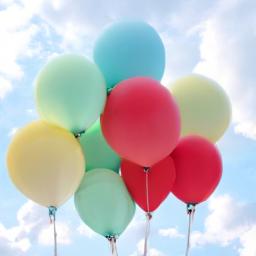}
  {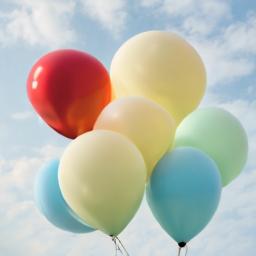}
  {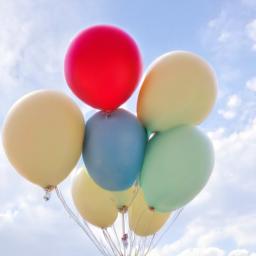}

\PromptPanelSeven
  {A sailboat drifts lazily along the river, with swans paddling gently beside it and willows weeping at its banks.}
  {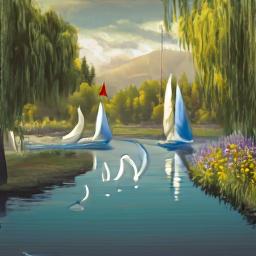}
  {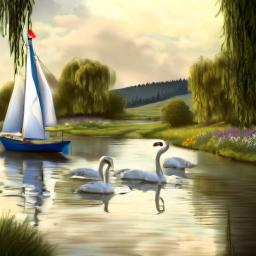}
  {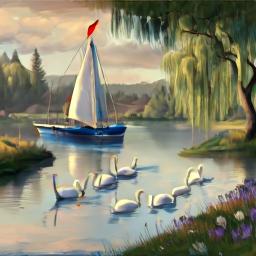}
  {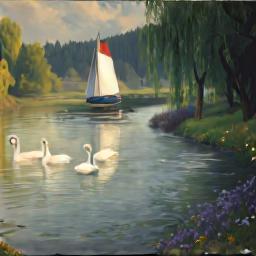}
  {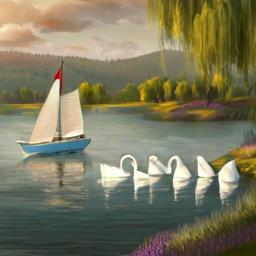}
  {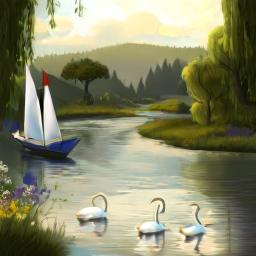}
  {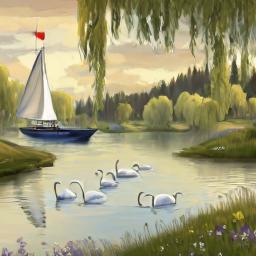}
  {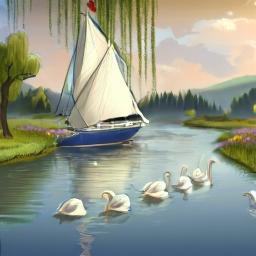}

\PromptPanelSeven
  {At the beach, three seashells lie close together: the largest has a spiraled pattern, the medium one is smooth and white, and the smallest has stripes.}
  {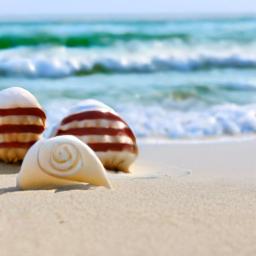}
  {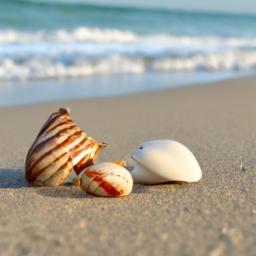}
  {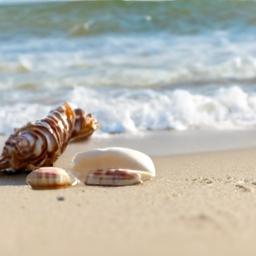}
  {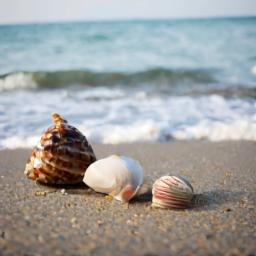}
  {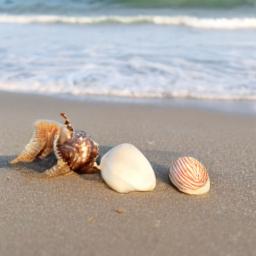}
  {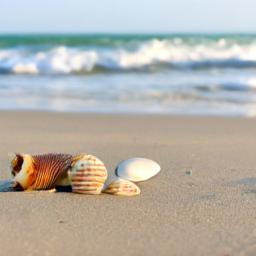}
  {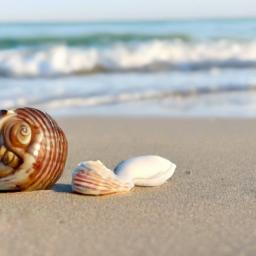}
  {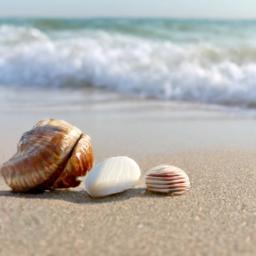}

\PromptPanelSeven
  {A mirror without a reflection of light.}
  {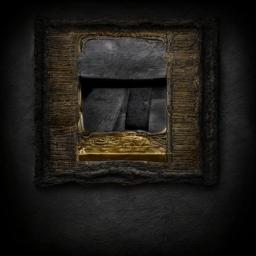}
  {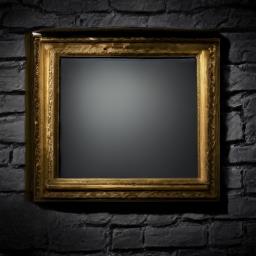}
  {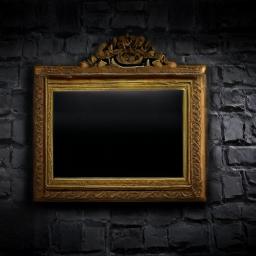}
  {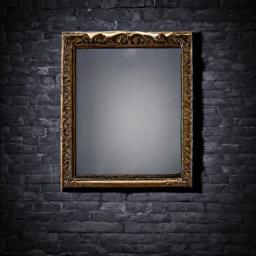}
  {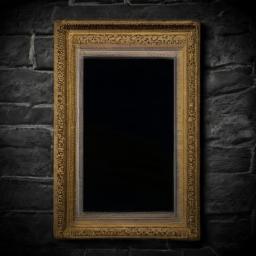}
  {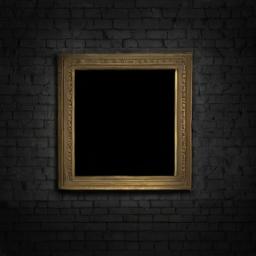}
  {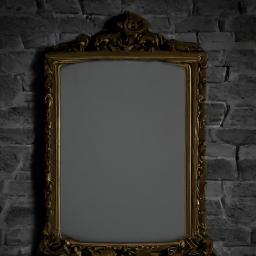}
  {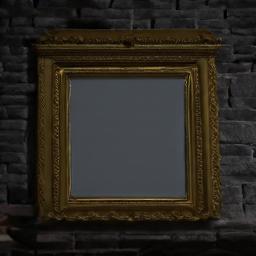}

\PromptPanelSeven
  {A calendar without any markings.}
  {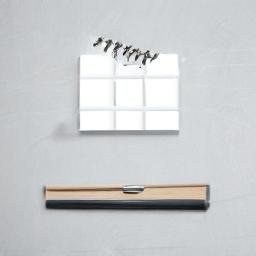}
  {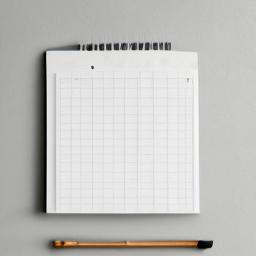}
  {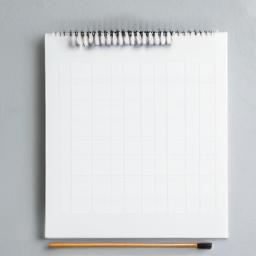}
  {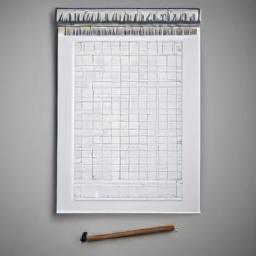}
  {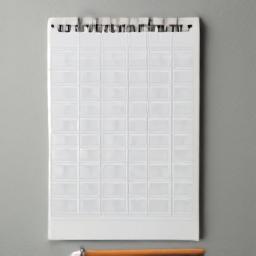}
  {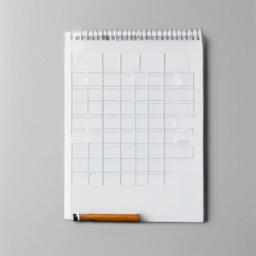}
  {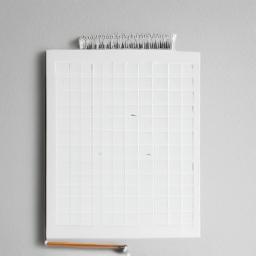}
  {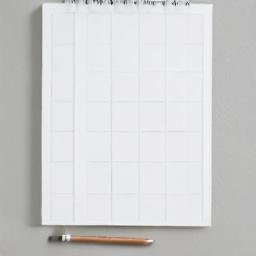}

\PromptPanelSeven
  {A painting where the mountain is depicted as taller than the trees in the foreground.}
  {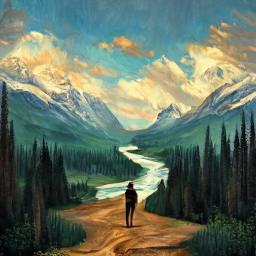}
  {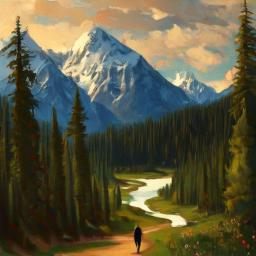}
  {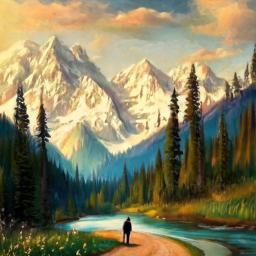}
  {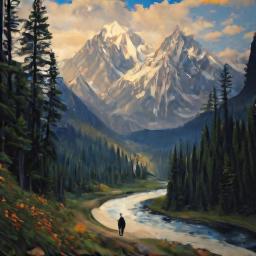}
  {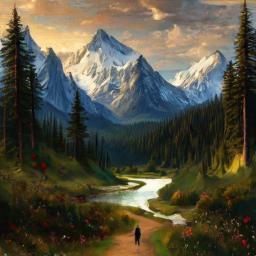}
  {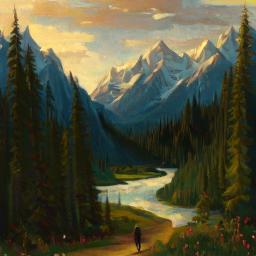}
  {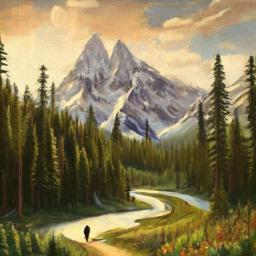}
  {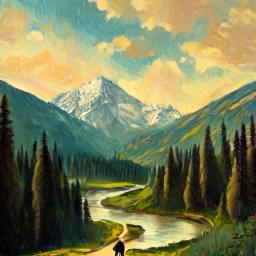}

\PromptPanelSeven
  {Six birds soar above two towering trees.}
  {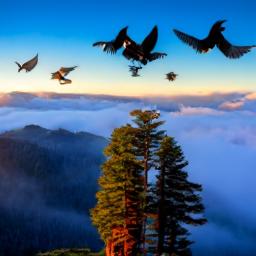}
  {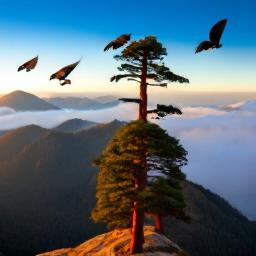}
  {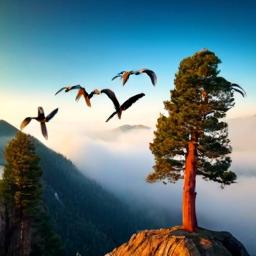}
  {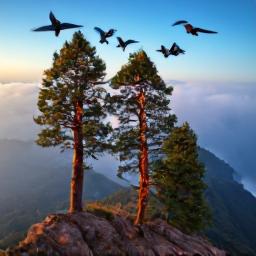}
  {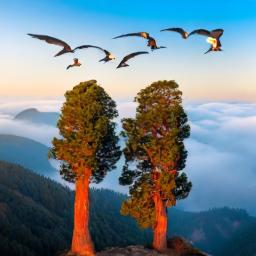}
  {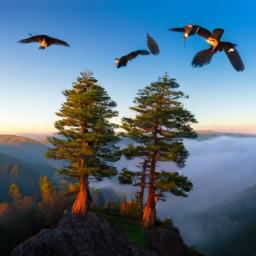}   
  {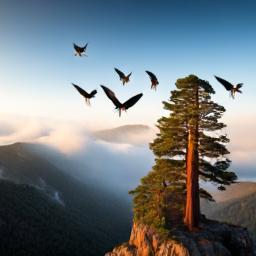}
  {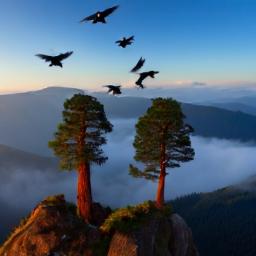}

\PromptPanelSeven
  {Four red apples in a basket on the kitchen counter.}
  {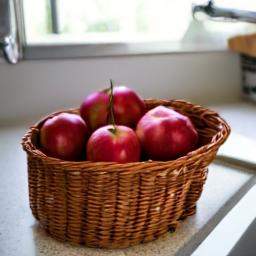}
  {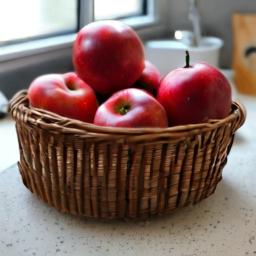}
  {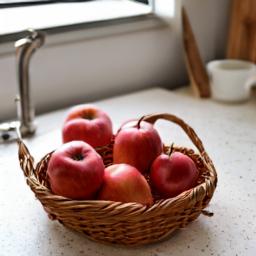}
  {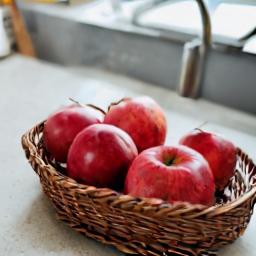}
  {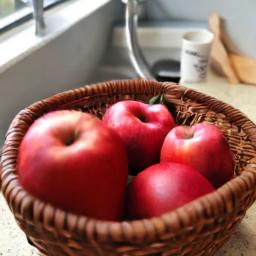} 
  {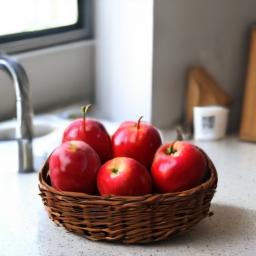}
  {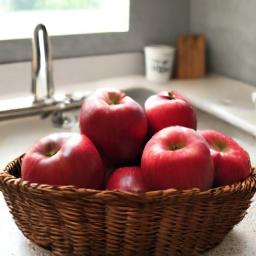}
  {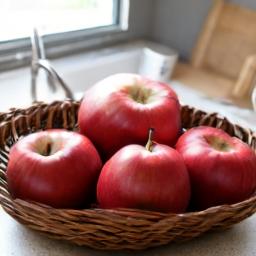}

\PromptPanelSeven
  {A vibrant blue rose blooming on a red lush rose bush.}
  {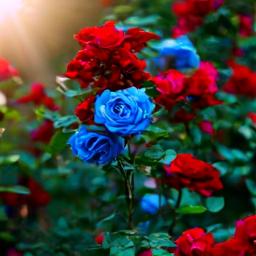}
  {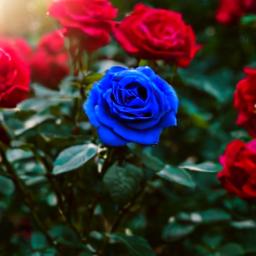}
  {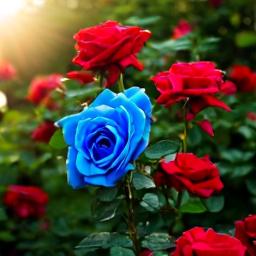}
  {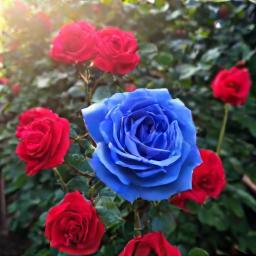}
  {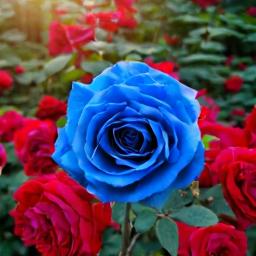} 
  {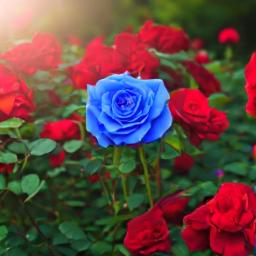}
  {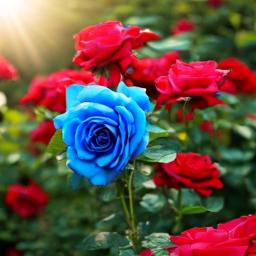}
  {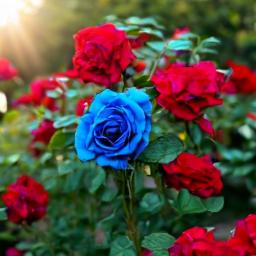}

\PromptPanelSeven
  {A tomato vine with several tomatoes on it, all yellow except the largest which is red.}
  {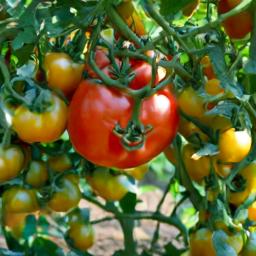}
  {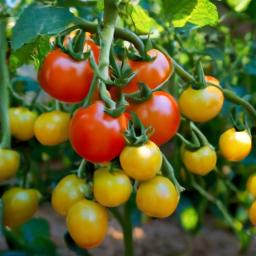}
  {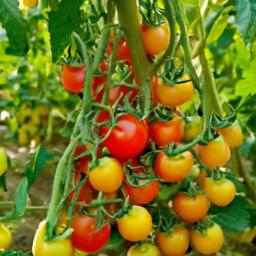}
  {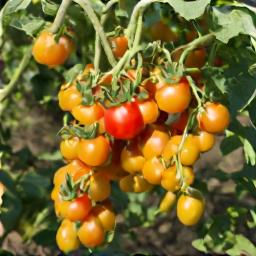}
  {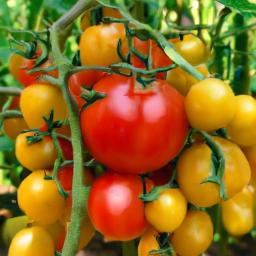} 
  {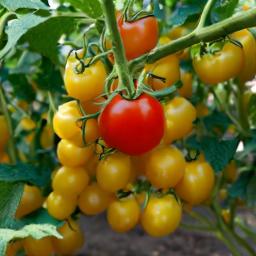}
  {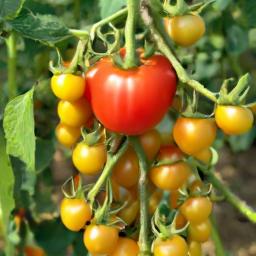}
  {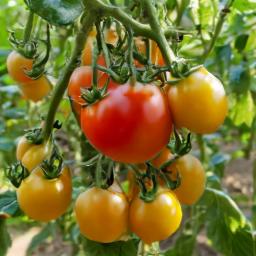}

\PromptPanelSeven
  {An oil painting of a majestic lion hanging above the fireplace.}
  {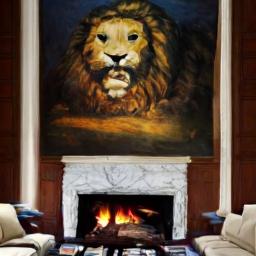}
  {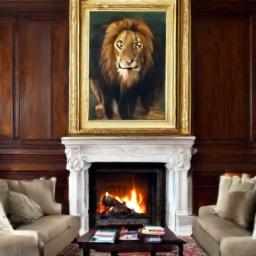}
  {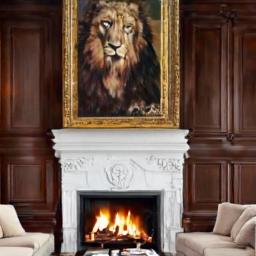}
  {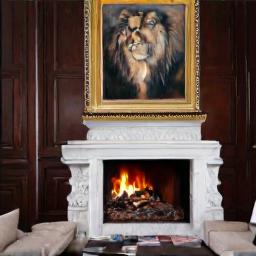}
  {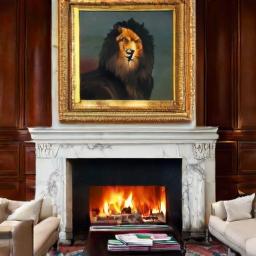}
  {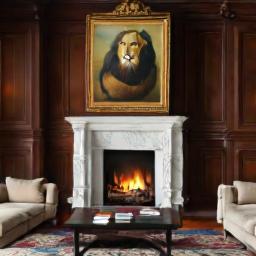}
  {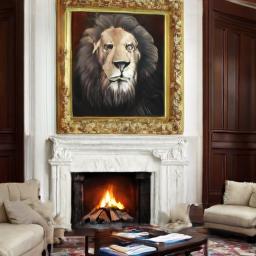}
  {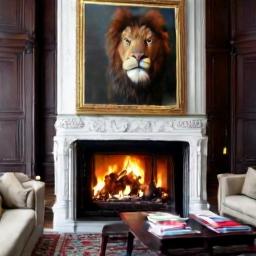}

\PromptPanelSeven
  {A serene stream winding through a lush meadow at sunrise.}
  {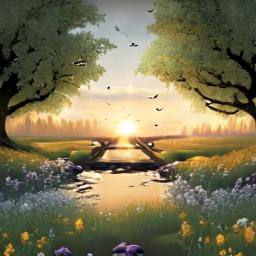}
  {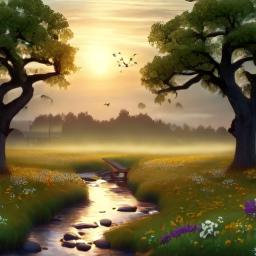}
  {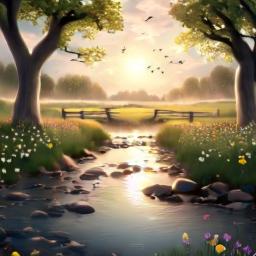}
  {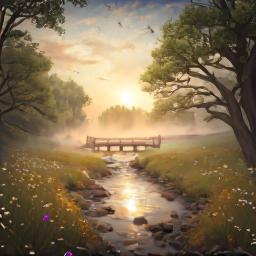}
  {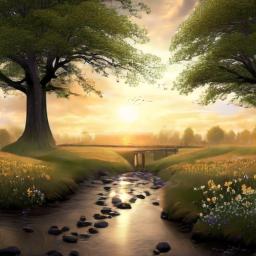}
  {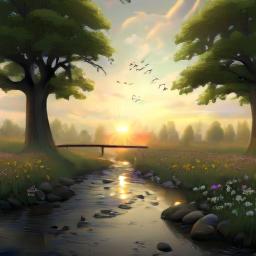}
  {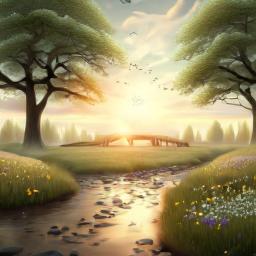}
  {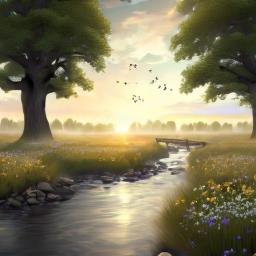}

\PromptPanelSeven
  {Three ceramic cups sit to the right of a wooden fork.}
  {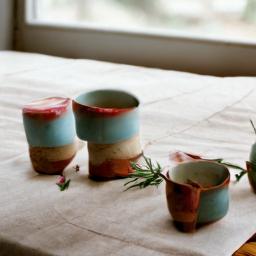}
  {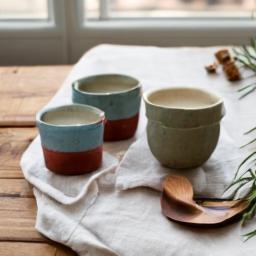}
  {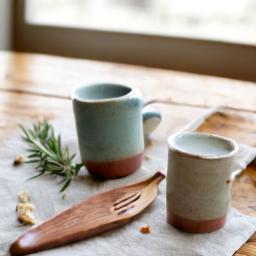}
  {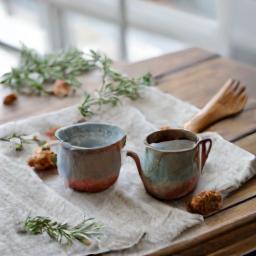}
  {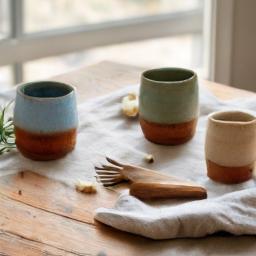}
  {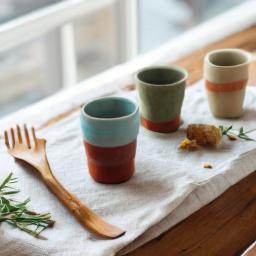}
  {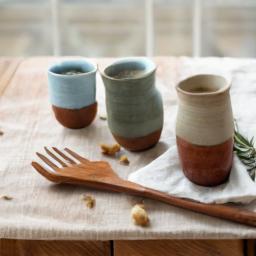}
  {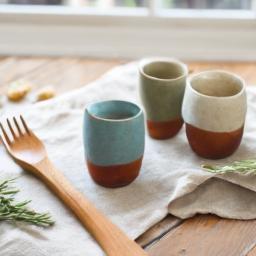}

\PromptPanelSeven
  {A vase contains flowers of various colors, but there are no red flowers.}
  {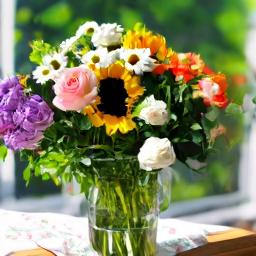}
  {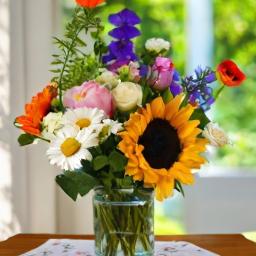}
  {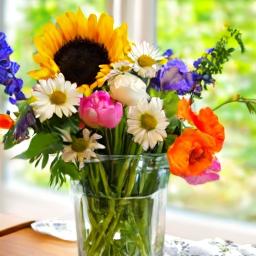}
  {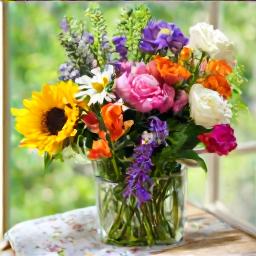}
  {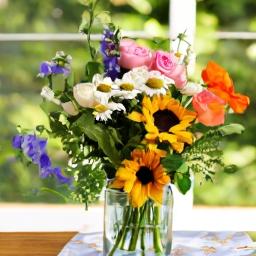}
  {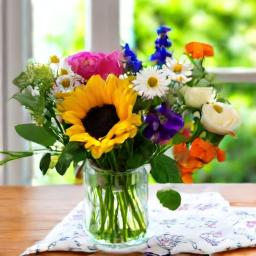}
  {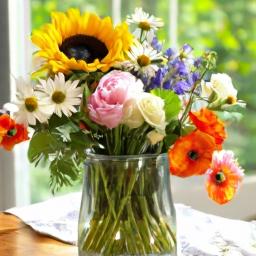}
  {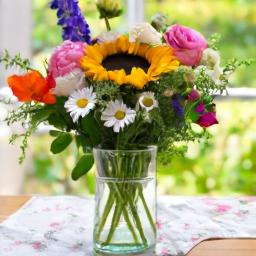}

\PromptPanelSeven
  {There is a glass without orange juice next to a glass filled with orange juice.}
  {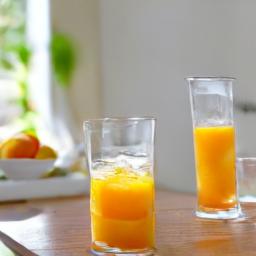}
  {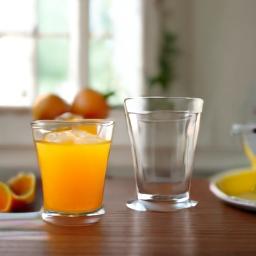}
  {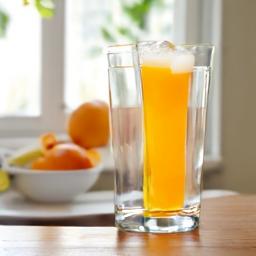}
  {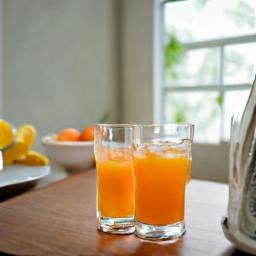}
  {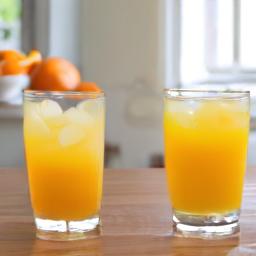}
  {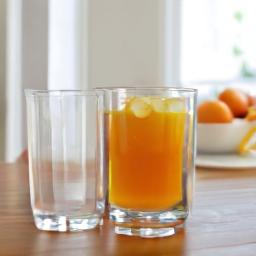}
  {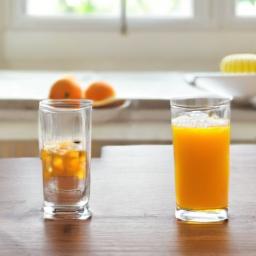}
  {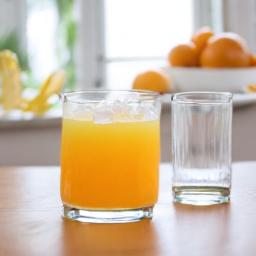}

\PromptPanelSeven
  {Three flowers in the meadow, with only the red rose blooming; the others are not open.}
  {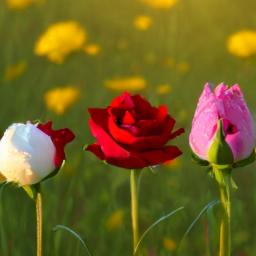}
  {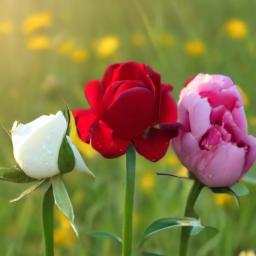}
  {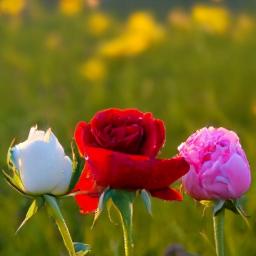}
  {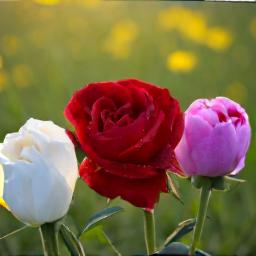}
  {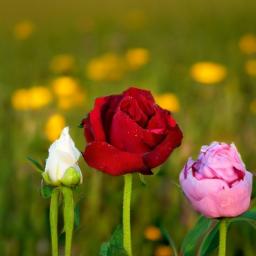}
  {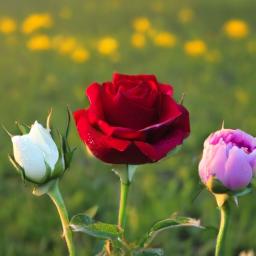}
  {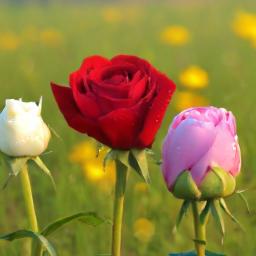}
  {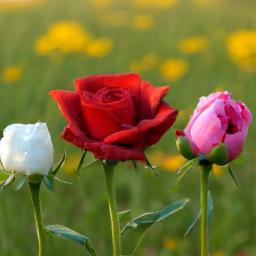}
  
\PromptPanelSeven
  {In a basket, there are many cauliflowers, all green except for the one in the middle, which is yellow.}
  {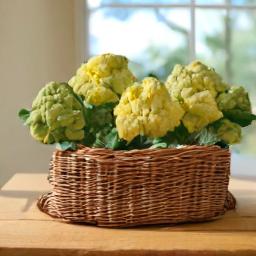}
  {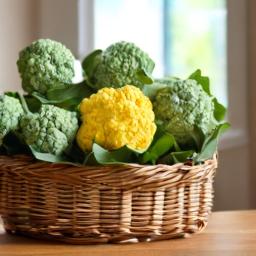}
  {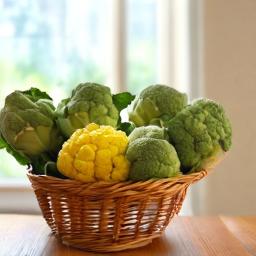}
  {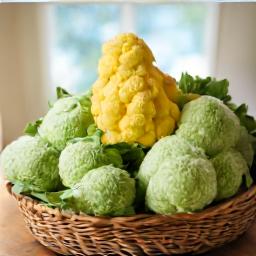} 
  {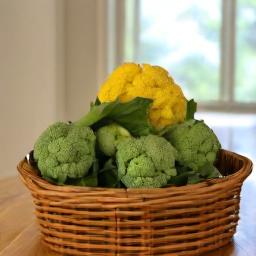}
  {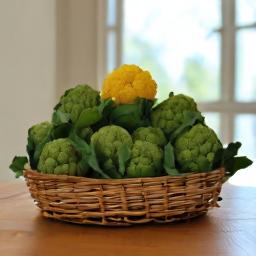}
  {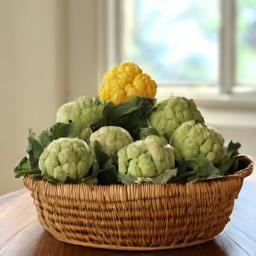}
  {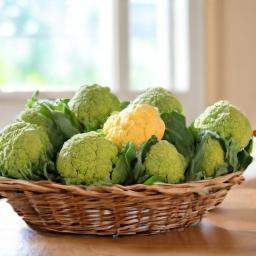}

\PromptPanelSeven
  {In the art gallery, two paintings hang on the wall, one a vibrant abstract with splashes of color, the other a detailed landscape with intricate brushwork.}
  {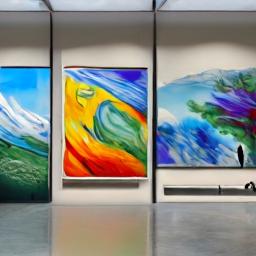}
  {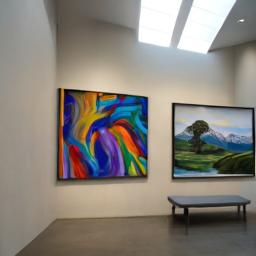}
  {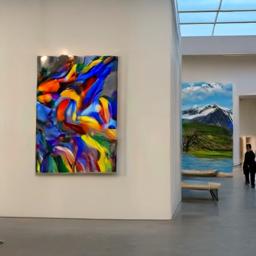}
  {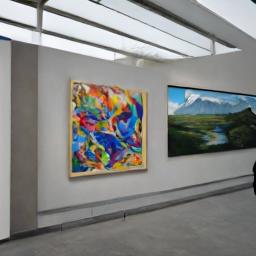}
  {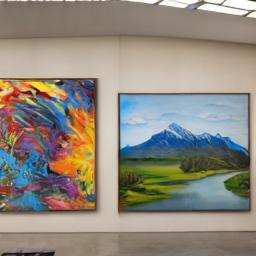}
  {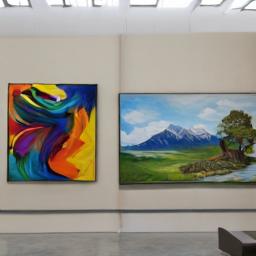}
  {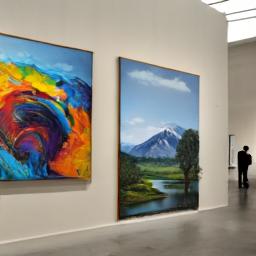}
  {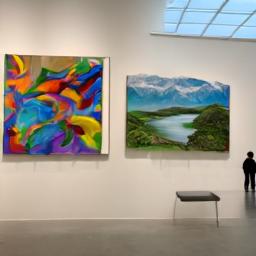}

\end{center}

\section{Experimental Details}
\label{sec:setup}
\subsection{Experimental Setup}
\begin{itemize}
    \item \textbf{Context Aggegator Setup:} For our text-only evaluation, the adapter module consists of a two-layer attention network with a hidden dimension of 1024. It is trained for one epoch on our 500k-sample dataset with a learning rate of 1e-5.
    \item \textbf{Text-to-Image Training:}  For the text-to-image task, we employ the Lumina-Image-2.0-2B ~\cite{qin2025lumina} architecture and train the models from scratch. For each text encoder under evaluation, we substitute it into the architecture while keeping all other components fixed. The models are trained at a 256x256 resolution on a private dataset, using a batch size of 512 for a total of 144k steps. In our two-step training strategy, we first train the layer weights $\omega_{i} $ jointly with the main model for 96k steps, after which we freeze $\omega_{i} $ for the remainder of the training.
    \item \textbf{Text-to-Video Training:} In contrast, for the text-to-video task, we fine-tune the pre-trained Wan2.1-T2V-1.3B model ~\cite{wan2025wan}. For each text encoder, we replace the original one and re-initialize the text embedding layers within the DiT module. To accelerate semantic alignment, we employ an interleaved training strategy where the ratio of training steps for image data (batch size 512) to video data (batch size 128) is $7:3$. The fine-tuning process consisted of a total of 20k training steps. In our two-step training strategy, we first train the layer weights $\omega_{i} $ jointly with the main model for 12k steps, after which we freeze $\omega_{i} $ for the remainder of the training.
\end{itemize}
In both the T2I and T2V training setups, the main backbone of the diffusion model remains fully trainable. The learning rate is set to 1e-4. For evaluation, we use GenAI-Bench ~\cite{li2024genai}.  During inference, all visual contents are generated using 50 denoising steps and the default negative prompt from GenAI-Bench. The CFG scale is set to 4.0 for the T2I task and 5.0 for the T2V task.

\subsection{Training Data Example}
This section presents concrete examples of the two key types of training data used in our experiments: (1)  the self-supervised contrastive learning data for training the Context Aggregator.and (2) the highly targeted VQA data for fine-tuning the GRAN-TED encoder.

\begin{figure}[ht]
\centering
   \includegraphics[width=0.5\textwidth, 
                 keepaspectratio, 
                 page=1]{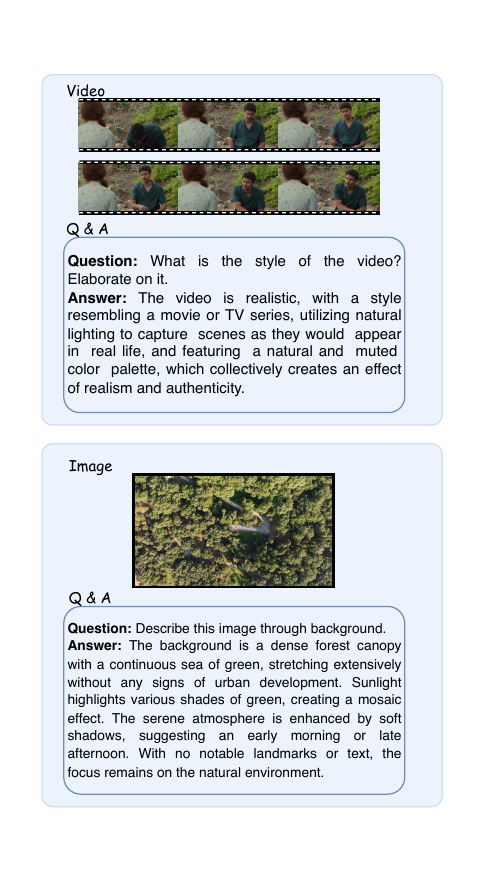} 
                 \caption{
                    Examples of visual question answering (VQA) training samples,
                    covering both image-based and video-based settings.
                  }
\label{fig:video_data}
\end{figure}

\subsubsection{Aggretator Training Data}

Our aggregator is trained using a contrastive learning paradigm, with the data construction process illustrated in Figure \cref{fig:cl_data}. For any given source caption, which serves as the Anchor, we generate a positive sample by prompting the model to re-caption the same visual content (image/video) using a different prompt. This yields a semantically similar but textually distinct caption. Concurrently, captions derived from entirely different visual content are treated as negative samples.

\begin{figure*}[t]
    \centering
    \begin{tcolorbox}[
        colback=gray!5,
        colframe=gray!60!black,
        title={Example Captions Used for Positive/Negative Statement Tasks},
        fonttitle=\bfseries,
        boxrule=0.5pt,
        arc=2pt
    ]
    \textbf{Caption (from Image/Video 1).}
  
    In a serene outdoor setting with bamboo and dim lighting, a middle-aged Asian man in traditional attire sits at a low wooden table. Wearing a white robe, a light brown vest, and a white headband, he is surrounded by a dimly lit courtyard or garden with bamboo trees and greenery. The scene, set at night, features a small wooden structure with a lantern and candles on the table. The man rises from his seat, turning his head with a serious and alert expression, hinting at concern. The realistic scene, possibly from a TV series or movie, emphasizes character portrayal with dramatic lighting, mixing natural and artificial sources to create a moody atmosphere. The muted color palette with earthy tones suggests a historical setting, while the camera captures the moment with a medium close-up shot, panning to the upper left at a normal speed. The lower angle positioning and shallow depth of field blur the distant background, transitioning from a view of the man's back to a profile.
  
    \medskip
    \textbf{Positive Caption (from Image/Video 1, different prompt).}
  
    A man of East Asian descent, likely in his 20s or 30s, with dark hair pulled back in a topknot and a white cloth draped over it, wears a light beige, loose-fitting robe with long sleeves and a white undergarment. He has a clean-shaven face, dark eyes, and an average build. In a serene outdoor setting at night, surrounded by bamboo and lanterns, he sits contemplatively before standing up with a serious expression. The scene, capturing a traditional or historical context, features soft, warm light from lanterns contrasting with dark greenery. Filmed with natural lighting, a muted color palette, and shallow depth of field, the medium close-up shot tilts upward, framing the man slightly to the right, evoking a somber, contemplative mood.
  
    \medskip
    \textbf{Negative Caption (from Image/Video 2).}
  
    A motorcyclist rides down a wide, empty road lined with trees and greenery on a sunny day. The motorcyclist, dressed in dark clothing, is positioned in the center of the frame, moving away from the camera. The road features clear lane lines, sidewalks, and is flanked by greenery and trees. The motorcycle is not clearly visible due to the distance. The background includes a grassy hillside on the right, with streetlights and trees along the road. The sky is clear and blue, creating a realistic, documentary-style scene with a natural, unedited feel. The motorcyclist continues straight, gradually becoming smaller in the frame before disappearing from view. The scene is characterized by medium saturation, moderate contrast, and brightness, with a normal color temperature. Captured with a deep depth of field, the camera pans left, maintaining a height roughly equal to the motorcyclist's line of sight, framing the subject slightly to the left of center.
    \end{tcolorbox}
  
    \caption{An example of the contrastive learning data used to train the context aggregator. For a given  caption, a Positive Sample is generated by re-captioning the same visual content (image/video) with a different prompt. Captions from entirely different visual content are used as Negative Samples. 
    }
    \label{fig:cl_data}
  \end{figure*}

\subsubsection{GRAN-TED Training Data}

Specific examples of the Visual Question Answering (VQA) data used for our fine-tuning, illustrating the video-based and image-based VQA pairs, are shown in Figure \cref{fig:video_data}.

\subsection{Prompt}

In this section, we provide the detailed prompts used throughout our experiments. Specifically, we include:
\begin{itemize}
    \item prompts for generating the base captions for images and videos
    (Figs.~\cref{fig:prompt-image-caption-1},
    \cref{fig:prompt-video-caption-1});
    \item prompts for constructing the positive and negative statements
    for the TED-6K benchmark (Figs.~\cref{fig:prompt-positive-statement},
    \cref{fig:prompt-negative-statement},
    together with the statement type definitions in
    Fig.~\cref{fig:statement-types-part1}) and Fig.~\cref{fig:statement-types-part2});
    \item prompts for generating the positive pairs required for contrastive
    learning      (Figs.~\cref{fig:prompt-image-caption-1},
    \cref{fig:prompt-image-caption-2},
    \cref{fig:prompt-video-caption-1},
    and~\cref{fig:prompt-video-caption-2}); 
\end{itemize}

\begin{figure*}[t]
    \centering
    \begin{tcolorbox}[
        colback=gray!5,
        colframe=gray!60!black,
        title={Prompt Used for Image Caption 1 Generation},
        fonttitle=\bfseries,
        boxrule=0.5pt,
        arc=2pt
    ]
    Your task is to generate a single, highly detailed (fine-grained), and coherent caption for the given image.
  
    \medskip
    \textbf{Core Principles}
    \begin{enumerate}[leftmargin=*]
      \item \textbf{Fidelity:} Your caption must be \emph{loyal to the visual facts} of the image. Do not invent details, assumptions, or information that is not clearly visible.
      \item \textbf{Coherence:} Your final output must be a \emph{single, flowing paragraph}. The description should be seamless and natural, prioritizing readability and flow.
    \end{enumerate}
  
    \textbf{Path to a Detailed Caption}
    Start by identifying the ``big picture'' (the main event or subject). Then, mentally review the following dimensions to see if you can add more detail naturally, without disrupting the narrative flow:
  
    \begin{itemize}[leftmargin=*]
      \item \textbf{The Scene (Basic Event):} What is the overall setting or event?
      \item \textbf{The ``Doing'' (Actions):} What are the key subjects (people, animals) doing?
      \item \textbf{The ``Placing'' (Spatial Relationships):} Where are things located in relation to each other?
      \item \textbf{The ``Attributes'' (Adjectives, Adverbs \& Quantity):}
        \begin{itemize}
          \item What are the qualities of subjects/objects? (e.g., color, texture, shape)
          \item How are actions being performed? (e.g., walking slowly)
          \item What is the degree of a quality? (e.g., very tall, incredibly bright)
          \item How many are there?
        \end{itemize}
      \item \textbf{The ``Clarity'' (Coreference):} Clearly link actions and descriptions to their correct subjects, ensuring there is no ambiguity about who is doing what.
      \item \textbf{The ``Text'' (OCR):} Is there any clearly readable text (on signs, shirts)?
    \end{itemize}
  
    \textbf{Important.} This list is a tool for consideration, not a checklist to be completed. You do not need to force every one of these points into your caption. The final, synthesized caption is all that matters. It must be loyal, fine-grained, and---above all---coherent.
    \end{tcolorbox}
    \caption{Prompt used for Image Caption 1 generation.}
    \label{fig:prompt-image-caption-1}
  \end{figure*}

  \begin{figure*}[t]
    \centering
    \begin{tcolorbox}[
        colback=gray!5,
        colframe=gray!60!black,
        title={Prompt Used for Image Caption 2 Generation},
        fonttitle=\bfseries,
        boxrule=0.5pt,
        arc=2pt
    ]
    Your task is to generate one high-quality, fine-grained caption for the provided image.
  
    \medskip
    \textbf{Core Objectives}
    \begin{enumerate}[leftmargin=*]
      \item \textbf{Fidelity (Be Factual):} Describe only what is visually present in the image. Do not invent or assume details, emotions, or relationships that are not explicitly shown.
      \item \textbf{Coherence (Be Natural):} The caption must be a single, well-written paragraph. It should be fluid and easy to read, not just a list of facts.
    \end{enumerate}
  
    \textbf{A Guide to Rich Detail}
    To ensure your caption is thorough, consider the following elements.
    This is a guide for inspiration, not a strict checklist; your primary goal is a natural, coherent narrative.
    Only include these details if they fit seamlessly.
  
    \begin{itemize}[leftmargin=*]
      \item \textbf{1. The Scene (Basic Event):} What is the setting? (e.g., a park, a kitchen, a concert)
      \item \textbf{2. Key Actions:} What are the subjects doing?
      \item \textbf{3. Visual Properties (Adjectives, Adverbs, Quantity):}
        \begin{itemize}
          \item What do things look like? (e.g., red, wooden, small)
          \item How are actions/qualities modified? (e.g., running quickly, very bright)
          \item How many key items are there?
        \end{itemize}
      \item \textbf{4. Layout (Spatial Relationships):} Where are things located in relation to each other? (e.g., on the table, next to the window)
      \item \textbf{5. Clarity (Coreference):} Ensure it is obvious which description belongs to which subject.
      \item \textbf{6. Visible Text (OCR):} Is there any text on signs, clothing, or objects that is clear enough to read?
    \end{itemize}
  
    Your final output is the caption itself. It must be a single, coherent paragraph that is both detailed and strictly faithful to the image.
    \end{tcolorbox}
  
    \caption{Prompt used for Image Caption 2 generation.}
    \label{fig:prompt-image-caption-2}
  \end{figure*}

  \begin{figure*}[t]
    \centering
    \begin{tcolorbox}[
        colback=gray!5,
        colframe=gray!60!black,
        title={Prompt Used for Video Caption 1 Generation},
        fonttitle=\bfseries,
        boxrule=0.5pt,
        arc=2pt
    ]
    Your task is to generate a single, highly detailed (fine-grained), and coherent caption for the given video.
  
    \medskip
    \textbf{Core Principles}
    \begin{enumerate}[leftmargin=*]
      \item \textbf{Fidelity:} Your caption must be loyal to the visual facts of the entire video clip. Do not invent details, assumptions, or information that is not clearly visible.
      \item \textbf{Coherence:} Your final output must be a single, flowing paragraph. The description should be seamless and natural, prioritizing readability and flow.
    \end{enumerate}
  
    \textbf{Path to a Detailed Caption}
    Start by identifying the big picture (the main event or subject). Then, mentally review the following dimensions to see if you can add more detail naturally, without disrupting the narrative flow:
  
    \begin{itemize}[leftmargin=*]
      \item \textbf{The Scene (Basic Event):} What is the overall setting or event?
      \item \textbf{The ``Doing'' (Actions):} What are the key subjects (people, animals) doing over time? (e.g., running, talking, opening)
      \item \textbf{The ``Timing'' (Temporal Relationship):} How do events unfold or relate in time?
      \item \textbf{The ``Placing'' (Spatial Relationships):} Where are things located in relation to each other, and do these relationships change?
      \item \textbf{The ``Attributes'' (Adjectives, Adverbs \& Quantity):}
        \begin{itemize}
          \item What are the qualities of subjects/objects? (e.g., color, texture, shape)
          \item How are actions being performed? (e.g., walking slowly)
          \item What is the degree of a quality? (e.g., very tall)
          \item How many are there?
        \end{itemize}
      \item \textbf{The ``Clarity'' (Coreference):} Clearly link actions and descriptions to their correct subjects.
      \item \textbf{The ``Text'' (OCR):} Is there any clearly readable text (on signs, shirts) that appears long enough to be read?
    \end{itemize}
  
    This list is a tool for consideration, not a checklist to be completed.
    The final, synthesized caption must be loyal, fine-grained, and coherent.
    \end{tcolorbox}
  
    \caption{Prompt used for Video Caption 1 generation.}
    \label{fig:prompt-video-caption-1}
  \end{figure*}

  \begin{figure*}[t]
    \centering
    \begin{tcolorbox}[
        colback=gray!5,
        colframe=gray!60!black,
        title={Prompt Used for Video Caption 2 Generation},
        fonttitle=\bfseries,
        boxrule=0.5pt,
        arc=2pt
    ]
    Your task is to generate one high-quality, fine-grained caption for the provided video.
  
    \medskip
    \textbf{Core Objectives}
    \begin{enumerate}[leftmargin=*]
      \item \textbf{Fidelity (Be Factual):} Describe only what is visually present across the duration of the video. Do not invent or assume details, emotions, or relationships that are not explicitly shown.
      \item \textbf{Coherence (Be Natural):} The caption must be a single, well-written paragraph. It should be fluid and easy to read, not just a list of facts.
    \end{enumerate}
  
    \textbf{A Guide to Rich Detail}
    To ensure your caption is thorough, consider the following elements.
    This is a guide for inspiration, not a strict checklist;
    only include these details if they fit seamlessly.
  
    \begin{itemize}[leftmargin=*]
      \item \textbf{1. The Scene (Basic Event):} What is the setting? (e.g., a park, a kitchen, a concert)
      \item \textbf{2. Key Actions:} What are the subjects doing? (e.g., running, talking, changing expression)
      \item \textbf{3. Event Flow (Temporal Relationship):} What is the sequence or timing of events?
      \item \textbf{4. Visual Properties (Adjectives, Adverbs, Quantity):}
        \begin{itemize}
          \item What do things look like? (e.g., red, wooden, small)
          \item How are actions/qualities modified? (e.g., running quickly, very bright)
          \item How many key items are there?
        \end{itemize}
      \item \textbf{5. Layout (Spatial Relationships):} Where are things located in relation to each other?
      \item \textbf{6. Clarity (Coreference):} Ensure it is obvious which description belongs to which subject.
      \item \textbf{7. Visible Text (OCR):} Is there any text on signs, clothing, or objects that is clear enough to read?
    \end{itemize}
  
    Your final output is the caption itself.
    It must be a single, coherent paragraph that is both detailed and strictly faithful to the video.
    \end{tcolorbox}
  
    \caption{Prompt used for Video Caption 2 generation.}
    \label{fig:prompt-video-caption-2}
  \end{figure*}

  \begin{figure*}[t]
    \centering
    \begin{tcolorbox}[
        colback=gray!5,
        colframe=gray!60!black,
        title={Prompt Used for Positive Statement Generation},
        fonttitle=\bfseries,
        boxrule=0.5pt,
        arc=2pt
    ]
    You are provided with a long, detailed caption and a ``Statement Type''.
    Your task is to generate a clear, coherent statement that addresses the specific Statement Type,
    based only on the information given in the caption.
  
    \medskip
    \textbf{Instructions}
    \begin{enumerate}[leftmargin=*]
      \item Read the specified Statement Type (e.g., Basic Event, Action, Quantity, Spatial Relationship).
      \item Carefully read the long, detailed caption to find the specific information that corresponds to this Statement Type.
      \item If the information is clearly present, generate a declarative statement that only describes that Statement Type using the caption's information, without adding irrelevant details.
      \item If the caption does not contain explicit information for this perspective, the \texttt{"statement"} value in the JSON output must be \texttt{null}.
    \end{enumerate}
  
    Important notes: generate a statement only if the information is clearly present; do not make assumptions.
    The Statement Type dictates the perspective.
    Use \texttt{null} only when there is a mismatch between the caption's content and the requested Statement Type.
  
    \medskip
    \textbf{Input}
  
    Statement Type: \texttt{\{statement\_type\_placeholder\}}\\
    Caption: \texttt{\{caption\_placeholder\}}
  
    \medskip
    \textbf{Output Format}
  
  \begin{verbatim}
  {
    "statement": "Your generated statement here.",
    "reasoning": "Brief explanation of how this statement was generated
                  from the specified 'Statement Type' perspective, OR
                  why it could not be."
  }
  \end{verbatim}
  
    \end{tcolorbox}
  
    \caption{Prompt used for positive statement generation.}
    \label{fig:prompt-positive-statement}
  \end{figure*}

  \begin{figure*}[t]
    \centering
    \begin{tcolorbox}[
        colback=gray!5,
        colframe=gray!60!black,
        title={Prompt Used for Negative Statement Generation},
        fonttitle=\bfseries,
        boxrule=0.5pt,
        arc=2pt
    ]
    You are provided with a long, detailed caption, a ``Statement Type'',
    and a ``Positive Statement'' that is factually true according to the caption.
    Your task is to generate exactly three Negative Statements that are plausible
    but factually incorrect according to the caption.
  
    \medskip
    \textbf{Inputs}
    \begin{itemize}[leftmargin=*]
      \item Caption: \texttt{\{caption\_placeholder\}}
      \item Statement Type: \texttt{\{statement\_type\_placeholder\}}
      \item Positive Statement: \texttt{\{positive\_statement\_placeholder\}}
    \end{itemize}
  
    \textbf{Requirements for Negative Statements}
    \begin{enumerate}[leftmargin=*]
      \item \textbf{Maintain Type:} Each negative statement must be of the same Statement Type as the positive.
      \item \textbf{Contradict Caption:} Each negative statement must be demonstrably false based on the information in the caption.
      \item \textbf{Be Plausible \& Confusing:} Negatives should be minimally different from the positive or be plausible alternatives that could have been true but are not, making them hard to distinguish from the positive statement without carefully reading the caption.
    \end{enumerate}
  
    \textbf{Output Format}
  
    Please output only a JSON object containing exactly three strings:
  
  \begin{verbatim}
  {
    "negative_statements": [
      "Your first negative statement here.",
      "Your second negative statement here.",
      "Your third negative statement here."
    ]
  }
  \end{verbatim}
  
    \end{tcolorbox}
  
    \caption{Prompt used for negative statement generation.}
    \label{fig:prompt-negative-statement}
  \end{figure*}

\begin{figure*}[t]
\centering
\begin{tcolorbox}[
    colback=gray!5,
    colframe=gray!60!black,
    title={Statement Types (Part 1/2)},
    fonttitle=\bfseries,
    boxrule=0.5pt,
    arc=2pt,
    fontupper=\small,
    boxsep=2pt,
    left=4pt, right=4pt, top=4pt, bottom=4pt
]
\textbf{1. Basic Event Statement}\\
\textit{Description:} A high-level summary of the main activity, occurrence, or overall situation depicted in the content. It captures the core event without going into fine-grained details.\\
\textit{Examples:} ``The content shows a family having a picnic in the park.''; ``A person is giving a presentation in a conference room.''; ``Two cats are playing in a living room.''

\medskip
\textbf{2. Action Recognition Statement}\\
\textit{Description:} Identifies and describes a specific, concrete action performed by a subject (person or animal). More granular than a Basic Event statement, focusing on a single action rather than the entire scene.\\
\textit{Examples:} ``A woman is chopping vegetables.''; ``The cat is stretching.''; ``A man opened a book.''

\medskip
\textbf{3. Temporal Statement}\\
\textit{Description:} Describes the temporal order or sequence of events within the content, using clear temporal connectors (e.g., ``first'', ``then'', ``after'', ``while'').\\
\textit{Examples:} ``First, the man opened the door, then he walked inside.''; ``The audience started clapping after the woman finished her speech.''; ``The dog was barking while the car was driving by.''

\medskip
\textbf{4. Spatial Relationship Statement}\\
\textit{Description:} Describes the spatial position and layout relationships between two or more objects or entities. Can be static (fixed positions) or dynamic (positions anchored to a specific time or event).\\
\textit{Examples:} ``The lamp is on the desk.''; ``The cat was under the chair when the man entered the room.''; ``The ball is to the left of the goal after it was kicked.''

\medskip
\textbf{5. Quantity Statement}\\
\textit{Description:} Explicitly identifies the exact number or count of specific objects, people, or entities.\\
\textit{Examples:} ``There are three people standing on the sidewalk.''; ``The content shows two red cars.''; ``One person is wearing a hat.''
\end{tcolorbox}

\caption{Definitions of the statement types used in our positive and negative statement generation tasks (Part 1: Types 1--5).}
\label{fig:statement-types-part1}
\end{figure*}

\begin{figure*}[t]
\centering
\begin{tcolorbox}[
    colback=gray!5,
    colframe=gray!60!black,
    title={Statement Types (Part 2/2)},
    fonttitle=\bfseries,
    boxrule=0.5pt,
    arc=2pt,
    fontupper=\small,
    boxsep=2pt,
    left=4pt, right=4pt, top=4pt, bottom=4pt
]
\textbf{6. Adjective Statement}\\
\textit{Description:} Describes a characteristic, quality, or state of being of a noun (object, person, etc.). The focus is the adjective itself.\\
\textit{Examples:} ``The car is red.''; ``A tall woman is visible.''; ``The cat is fluffy.''

\medskip
\textbf{7. Adverb Statement}\\
\textit{Description:} Describes the manner, place, time, frequency, or degree of an action or quality, with the adverb as the core.\\
\textit{Examples:} ``The man is walking slowly.''; ``The building is very tall.''; ``The light is flashing repeatedly.''

\medskip
\textbf{8. Coreference Statement}\\
\textit{Description:} Resolves an ambiguous reference by explicitly connecting it to its antecedent (e.g., linking a pronoun or descriptive noun phrase to the entity it refers to).\\
\textit{Examples:} ``The pronoun 'it' in the phrase '...a dog. It is barking.' refers to the dog.''; ``The phrase 'the main character' refers to the young white woman wearing a blue tank top.''

\medskip
\textbf{9. OCR Statement}\\
\textit{Description:} Transcribes clearly visible text content (OCR) within the content, directly quoting the text.\\
\textit{Examples:} ``The text on the sign reads 'STOP'.''; ``The person's shirt has the word 'University' on it.''; ``A bus with the destination 'Downtown' is visible.''
\end{tcolorbox}

\caption{Definitions of the statement types used in our positive and negative statement generation tasks (Part 2: Types 6--9).}
\label{fig:statement-types-part2}
\end{figure*}

\end{document}